\documentclass[10pt]{article}
\usepackage{graphicx}
\usepackage{amssymb,amsfonts,amsmath}
\graphicspath{{images/},{figures/},{images/trajects/}}
\usepackage[utf8x]{inputenc}
\usepackage[russian,spanish,english]{babel}
\usepackage[T1,T2A]{fontenc}
\usepackage[tight]{subfigure}
\renewcommand{\thesubfigure}{\alph{subfigure}} 
\usepackage{lscape}
\newcommand{\rme}{\mathrm{e}}
\newcommand{\rmd}{\mathrm{d}}
\newcommand{\eref}[1]{equation~\ref{#1}} \newcommand{\sref}[1]{section~\ref{#1}}
\newcommand{\fref}[1]{figure~\ref{#1}}  \newcommand{\tref}[1]{table~\ref{#1}}
\newcommand{\Fref}[1]{Figure~\ref{#1}}  \newcommand{\Tref}[1]{Table~\ref{#1}}

\newcommand{\figura}[1]{#1}

\usepackage{color}
\definecolor{colora}{rgb}{0.898438, 0.621094, 0.}
\definecolor{colorb}{rgb}{0.335938, 0.703125, 0.910156}
\definecolor{colorc}{rgb}{0., 0.617188, 0.449219}
\definecolor{colord}{rgb}{0.9375, 0.890625, 0.257813}
\definecolor{colore}{rgb}{0., 0.445313, 0.695313}
\definecolor{colorf}{rgb}{0.832031, 0.367188, 0.}

\usepackage{float}

\usepackage{cite}

\usepackage{hyperref}

\usepackage{lineno}

\usepackage{microtype}
\DisableLigatures[f]{encoding = *, family = * }

\usepackage[labelfont=bf,labelsep=period,justification=raggedright]{caption}

\makeatletter
\renewcommand{\@biblabel}[1]{\quad#1.}
\makeatother

\date{}

\begin{document}

\newcommand{\unam}{Universidad Nacional Aut\'onoma de M\'exico}
\newcommand{\ifunam}{Instituto de F\'isica, \unam}
\newcommand{\fc}{Facultad de Ciencias, \unam}
\newcommand{\ccc}{Centro de Ciencias de la Complejidad, \unam}
\newcommand{\iimas}{Instituto de Investigaciones en Matem\'aticas Aplicadas y en Sistemas, \unam}

% Title must be 150 characters or less
\begin{flushleft}
{\Large
\textbf{\textcolor{black}{Rank diversity of languages: Generic behavior in computational linguistics}}
}
% Insert Author names, affiliations and corresponding author email.

Germinal Cocho$^{1,2}$, Jorge Flores$^{1}$, Carlos Gershenson$^{3,2,\ast}$, 
Carlos Pineda$^{1}$, Sergio S\'anchez$^{4}$
\\
\bf{1} \ifunam
\\
\bf{2} \ccc
\\
\bf{3} \iimas
\\
\bf{\textcolor{black}{4}} \fc
\\
$\ast$ E-mail: cgg@unam.mx
\end{flushleft}

% Please keep the abstract between 250 and 300 words
\section*{Abstract}

\textcolor{black}{
Statistical studies of languages have focused on the rank-frequency distribution of words. Instead, we introduce here a measure of how word ranks change in time and call this distribution \emph{rank diversity}. We calculate this diversity for books published in six European languages since 1800, and find that it follows a universal lognormal distribution. Based on the mean and standard deviation associated with the lognormal distribution, we define three different word regimes of languages: ``heads'' consist of words which almost do not change their rank in time, ``bodies'' are words of general use, while ``tails'' are comprised by context-specific words and vary their rank considerably in time. The heads and bodies reflect the size of language cores identified by linguists for basic communication. We propose a Gaussian random walk model which reproduces the rank variation of words in time and thus the diversity. Rank diversity of words can be understood as the result of random variations in rank, where the size of the variation depends on the rank itself. We find that the core size is similar for all languages studied.
}
\section*{Introduction}

Statistical studies of languages have become popular since the work
of George Zipf~\cite{zipf} and have been refined with the availability of large
data sets and the \textcolor{black}{introduction} of novel analytical
models~\cite{mandelbrot1953informational, hawkins1992evolution, %165464,
ZipfRnd2002, %Ha02extensionof, FerreriCancho04022003, 
1367-2630-13-4-043004,
PhysRevE.83.036115, Perc07122012}.  Zipf found that when words of large corpora
are ranked according to their frequency, there seems to be a universal tendency
across texts and languages. He proposed that ranked words follow a power law $f\sim1/k$, where
$k$ is the rank of the word---the higher ranks corresponding to the least
frequent words---and $f$ is the relative frequency of each
word~\cite{newman2005power,clauset2009power}. This regularity of languages and
other social and physical
phenomena had been noticed beforehand, at least by Jean-Baptiste
Estoup~\cite{Petruszewycz1973Lhistoire-de-la} and Felix
Auerbach~\cite{Auerbach1913}, but it is now known as Zipf's law. 

Zipf's law is a rough approximation \textcolor{black}{of} the precise statistics of
rank-frequency distributions of languages. As a consequence, several variations
have been proposed~\cite{Booth1967386, Montemurro2001567,
1367-2630-15-9-093033, PhysRevX.3.021006}.  We compared Zipf's law with four
other models, all of them behaving as $1/k^a$ for a small $k$, with $a \approx
1$, as detailed in the SI. We found that all models have systematic errors so
it was difficult to choose one over the other. 

Studies based on rank-frequency distributions of languages have proposed two
word regimes~\cite{PhysRevX.3.021006,FerrericanchoSole2001}: a ``core'' where
the most common words occur, which behaves as $1/k^a$ for small $k$, and
another region for large $k$, which is identified by a change of exponent $a$
in the distribution fit. Unfortunately, the point where exponent $a$ changes
varies widely across texts and languages, from
5000~\cite{FerrericanchoSole2001} to 62,000~\cite{PhysRevX.3.021006}.
\textcolor{black}{A recent study~\cite{Bochkarev2014Universals-vers} measures the number of most frequent words which account for $75\%$ of the Google books corpus. Differences of an order of magnitude across languages were obtained, from $2365$ to $21077$ words \textcolor{black}{(including inflections of the same stems)}. This illustrates the variability of rank-frequency distributions.}
 \textcolor{black}{The core of human languages can be considered }to be between 1500 and \textcolor{black}{3000}
words \textcolor{black}{(not counting different inflections of the same stems)}, based on basic vocabularies for
foreigners~\cite{Takala1985Estimating-stud}, creole~\cite{Hall-1953}, and
pidgin languages~\cite{romaine1988pidgin}. For example, Voice of America's
Special English~\cite{Beare2014Voice-of-Americ} and Wikipedia in Simple English
use about 1500 and 2000 words, respectively \textcolor{black}{(not counting inflections)}. The Oxford Advanced Learner's Dictionary lists 3000 priority \textcolor{black}{lexical entries}~\cite{hornby2005oxford}.
This suggests that \textcolor{black}{the change of exponent $a$ or another arbitrary cutoff in} rank-frequency
distributions does not reflect the size of the core of languages.

In view of these problems with rank-frequency distributions, we propose a
novel measure to characterize statistical properties of languages. We have called
this measure \emph{rank diversity} and it tells us how words change their rank
in time. With rank diversity, three regimes \textcolor{black}{of} words \textcolor{black}{are 
identified}: ``heads'', ``bodies'' and ``tails''. 
\textcolor{black}{
This measure of rank diversity
follows the same simple functional law with similar parameters
 for all data analyzed.}
% This measure of rank diversity
% is almost the same for all data analyzed; i
In particular, this is so for the six European languages studied 
\textcolor{black}{here} using a large
data set of more
than 6.4 $\times10^{11}$ words from Google Books~\cite{Michel14012011}, which
contains \textcolor{black}{about} $4\%$ of all books \textcolor{black}{written until 2008}. \textcolor{black}{It should be noted that this data set includes all different inflected forms (such as plural, different tense/aspect forms, etc.) found in the book corpus.} Data sets such as this have allowed
the study of ``culturomics'': how cultural traits such as language have changed
in time~\cite{%dodds2011temporal, 
wijaya2011understanding,
%petersen2012languages, 
serra2012measuring, petersen2012statistical,
blumm2012dynamics, 10.1371/journal.pone.0059030, perc2013self, %CPLX:CPLX21436,
Febres2013Complexity-meas}.

%{\bf creo qeu aca como estamos en un tipo introduccion podemos agregar que proponemos un modelo y encontramos muchas coincidencias. Mi propuesta.}

The rank diversity follows a \textcolor{black}{scale-invariant} behavior regarding its fluctuations,
which inspires a model based on random walks, with \textcolor{black}{scale-invariant} random
steps. This model reproduces the behavior of diversity and thus captures the
essence of the evolution of word rank across different languages. 
% }}}
\section*{Rank diversity \textcolor{black}{of} words} %{{{

In what follows we shall consider six European languages from the Indo-European
family. They are English and German; Spanish, French and Italian; and Russian.
They belong to different linguistic branches: Germanic, Romance, and Slavic,
respectively. The native speakers of these languages account for approximately 
17\% of the
world population.

%http://en.wikipedia.org/wiki/List_of_languages_by_number_of_native_speakers
%

We shall start by taking into account the 20, say, most used words in the six
languages, that is, the \textcolor{black}{lowest-ranked} words. Using, for the sake of uniformity,
\textcolor{black}{the first sense or first meaning} given by Google Translate, once these words are translated into
English, the coincidences in all six languages are remarkable (see Table S1
in File S1). This could have been
foreseen, since most of the \textcolor{black}{lowest-ranked} words are articles, prepositions or
conjunctions, \emph{i.e.} what \textcolor{black}{is} called function words. A different matter,
as we shall see, would result if we had considered only nouns, verbs, adverbs
or adjectives, known as content words. 

%\texttt{Germi: incluir figura de espagueti para las 20 primeras palabras}

\begin{figure}[H] % {{{ %%%Fig 1 Overlap of functional and content words
    \centerline{\figura{\includegraphics[width=.94\textwidth]{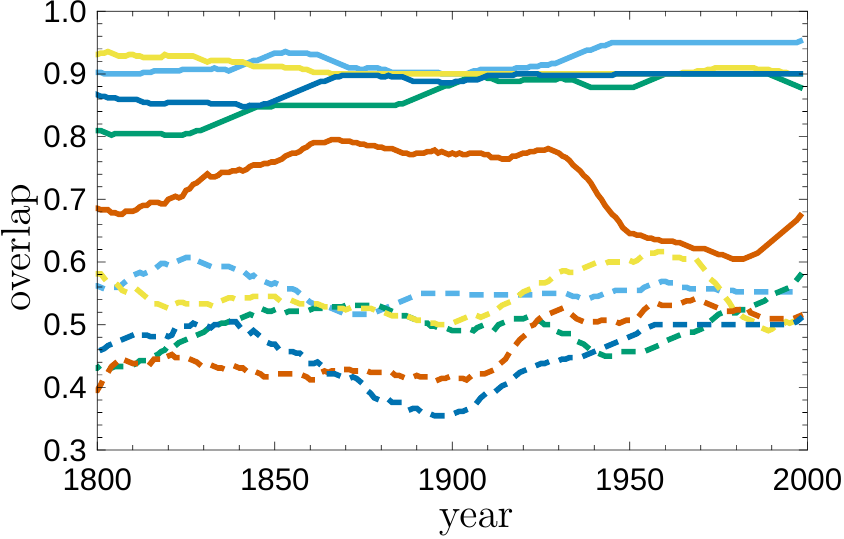} }}
     \caption{ 
     {\bf Overlap of the 20 most frequent
     words (continuous lines), and of the 20 most frequent
     \emph{content} words (dashed lines)} across languages, 
     with respect to English, as a function of time. When
     words have more than one meaning, the first sense, 
     according to Google
     Translate, was used.
     The color code for languages is as follows: light blue for French, green for German, yellow for Italian, dark blue for Spanish, and dark orange for Russian. Additionally, light orange will
be used for English when required (see also Fig.~\ref{fig:diversity}).  The same color coding for languages will be used throughout the rest of the article.  
     }
  \label{fig:overlap:evolution}
\end{figure}

In order to quantify this fact, we present in Fig.~\ref{fig:overlap:evolution}
the time evolution of the overlap of the first 20 \textcolor{black}{lowest-ranked} words in the five languages with respect to the corresponding list of English.
\textcolor{black}{From the upper part of this figure we can see that along two
centuries this overlap fluctuates around 0.9, a rather large number, except for
Russian, since this language does not have articles.} \textcolor{black}{These
data reveal that these Indo-European languages have shared structural
properties}, notwithstanding that they belong to distinct linguistic branches.

 % }}}

The \textcolor{black}{lowest-ranked} words used to construct the upper part of
Fig.~\ref{fig:overlap:evolution} are essentially the same along centuries (See
Figs S3-S8 in SI). But this is not the case for content words, as can be seen
in Table S2 in File S1. \textcolor{black}{First, and as also shown by the dashed
curves in Fig.~\ref{fig:overlap:evolution}, the overlap of these words with
respect to English for the other five languages (including Russian) is of the
order of 0.5.} These values are much lower than the overlap of function words.
Second, the most common nouns vary considerably with time.  On the one hand,
nouns like \textit{time}, \textit{man}, \textit{life} and their translation to
the other languages are present independently of the century. On the other
hand, words like \textit{god} and \textit{king} have a low rank in the
eighteenth century but have a larger rank in the last century.  The \textcolor{black}{rank change in time} of these nouns reflect cultural facts. 

\begin{figure}[H] % {{{ Diversidad
     \centering
%           \label{fig:d:data}
         \figura{\includegraphics[width=\textwidth]{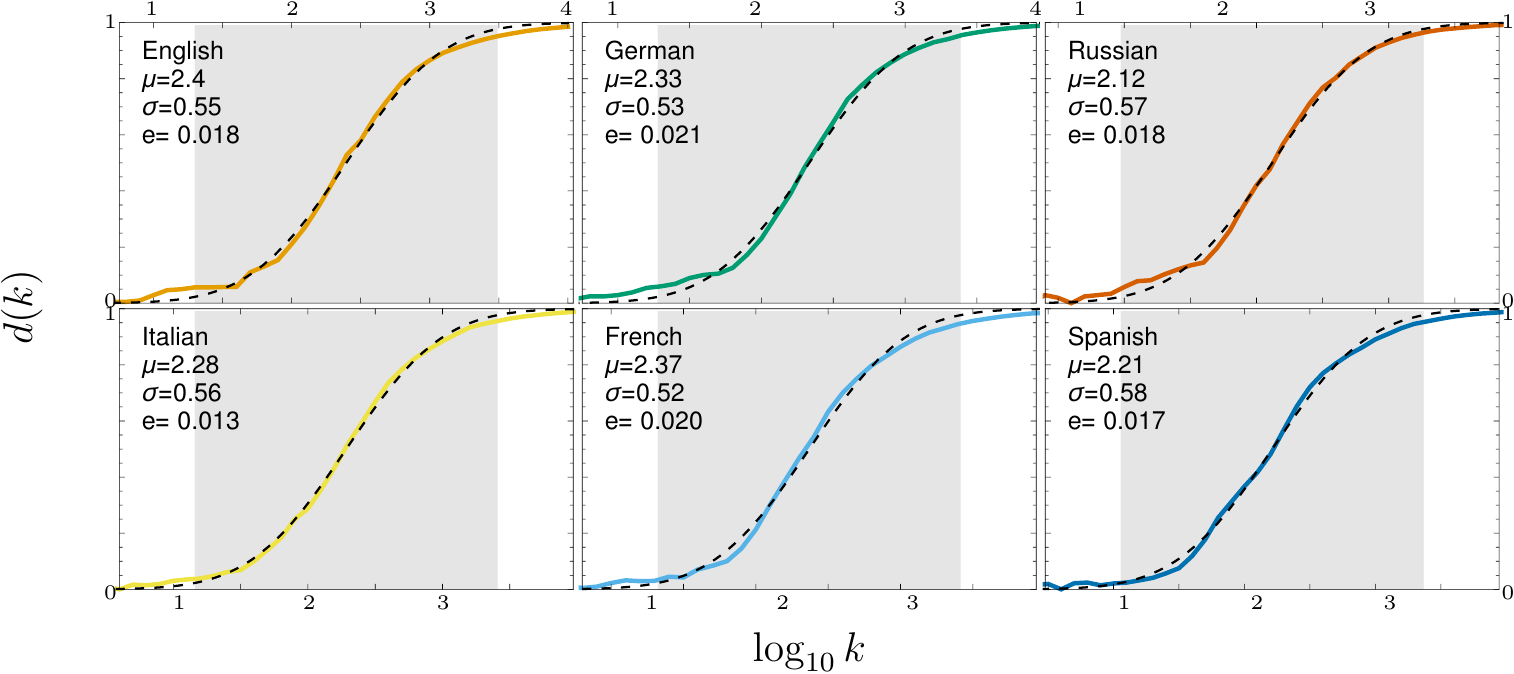}}
     \caption{{\bf Rank diversity.} Diversity $d$ as a function of the rank $k$
     for different languages from 1800 to 2008\textcolor{black}{, where $d(k)$
     measures how many different words appear for a given rank $k$ during the
     time considered} ($\Delta t = 208$).
     \textcolor{black}{For example, for English, $d(1)=1/208$, as the word `the'
     appears in the first rank for all years considered. Although we have analyzed
     up to $k\approx 10^6$, rank diversity for
     $k>10^4$ is not shown as $d(k) \approx 1$, \emph{i.e.}, a different word
     appears in each rank every year.}  Data are windowed over time, with a
     slot of size \textcolor{black}{$\delta \log_{10} k =0.1$}, for the sake of
     clearness. \textcolor{black}{Additionally, the sigmoid defined in
     \eref{eq:sigmoid} is shown as a black dashed curve, with the best fit
     parameters, also reported in each subfigure. The mean square  error $e$
     between the data and the fit, is also given.} The shaded region
     corresponds to the average ``body'' of all languages.
     }
  \label{fig:diversity}
\end{figure}

What is discussed in the previous paragraph is an example of what could be
called rank diversity $d(k)$.  This is, in the present study, the number of
different words occurring at a specific rank $k$ over a given period of time
$\Delta t$. We found that the resulting rank diversity curves for the six
languages studied between 1800 and 2008 \textcolor{black}{are} similar to each other, as
shown in Figs.~\ref{fig:diversity} and \ref{fig:diversity:normalized}.  Low ranks have a very low diversity, as
\textcolor{black}{few words} appear in the same ranks for the years we have
studied. 

%%%Fig 2

 % }}}

As shown by the continuous lines in Fig. \ref{fig:diversity}, the sigmoid curve
fits very well $d(k)$ for all languages considered, \textcolor{black}{except for
low $k$ where the statistical fluctuations are larger due to the small sample
size. } The sigmoid is the cumulative
of a Gaussian distribution, i.e. 
\begin{equation}
\Phi_{\mu,\sigma}(\log_{10} k)=\frac{1}{\sigma\sqrt{2\pi}}
    \int_{-\infty}^{\log_{10} k} \rme^{-\frac{(y-\mu)^2}{2\sigma^2}} {\rm d}y, 
\label{eq:sigmoid}
\end{equation}
and is given as a function of $\log k$.
% 
% As shown by the continuous lines in Fig. \ref{fig:d:data}, the sigmoid curve (the cumulative
% of a Gaussian distribution), as a function of $\log k$
% \begin{equation}
% \Phi_{\mu,\sigma}(\log_{10} k)=\frac{1}{\sigma\sqrt{2\pi}}
%     \int_{-\infty}^{\log_{10} k} \rme^{-\frac{(y-\mu)^2}{2\sigma^2}} {\rm d}y
% \label{eq:sigmoid}
% \end{equation}
% fits very well $d(k)$ for all languages considered, \textcolor{black}{except for low $k$ where the statistical
% fluctuations are larger due to the small sample size. }
% , with the mean $\mu=2.29$ and the width $\sigma=0.55$. 
The values of $\mu$ and $\sigma$ reported in  Fig. \ref{fig:diversity}  were
obtained adjusting \eref{eq:sigmoid} \textcolor{black}{to the rank diversity calculated for each 
individual language.}
% The values of $\mu$ and $\sigma$ were obtained adjusting \eref{eq:sigmoid} to
% the diversity averaged over the six languages. 
The mean value $\mu$ identifies the point where $d(k)\approx 0.5$, while
the width $\sigma$ gives the scale in which $d(k)$ gets close to its
extremal values. 
\textcolor{black}{When $\log k$ is much larger than 
$\mu +  \sigma$,  $\Phi_{\mu,\sigma}(\log k)$ gets exponentially close to one, whereas when $\log k$
is much smaller than $\mu -\sigma$ it gets exponentially close to zero. 
It is customary in statistics to define a bulk of the Gaussian 
between $\mu \pm 2 \sigma$, where 95\% of the population lies. Along the same lines, 
we define three regions, marked by 
\begin{equation}
\log_{10} k_{\pm}=\mu\pm2\sigma.
\label{eq:kpm}
\end{equation}
% The value of 2 is chosen, thus, arbitrarily, but with an eye in the 
% natural scale of the problem, namely, $\sigma$.
}
% 
% 
% 
% 
% 
% 
% In Fig.~\ref{fig:d:data} three regions are apparent. They are determined by the
% points
% \begin{equation}
% \log_{10} k_{\pm}=\mu\pm2\sigma.
% \label{eq:kpm}
% \end{equation}

\begin{figure}[H] % {{{ \mu (t), k+ y k-
     \centerline{\figura{\includegraphics[width=.5\textwidth]{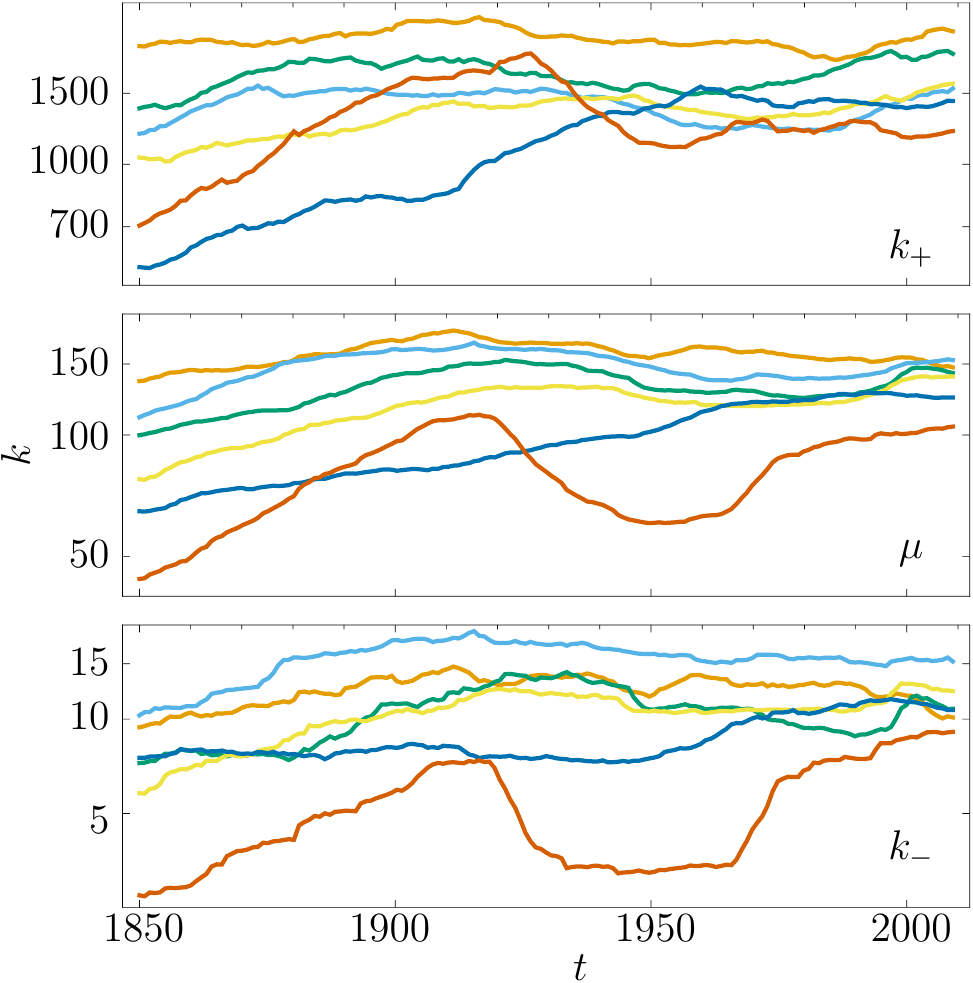}} }
     \caption{\textcolor{black}{{\bf Evolution in time of the center of the sigmoid (middle panel), and the
     borders of the head and body (bottom panel) and body and tail (top panel) for the different languages
     along time for intervals of fifty years, \emph{i.e. }$\Delta t = 50$.} Head
     words have $k \leq k_-$, body words have $k_- < k \leq k_+$, and tail
     words have $k_+ < k$.  See Fig.~\ref{fig:diversity} for color
     coding.}}
  \label{fig:mu:t}
\end{figure}

First, we find what we shall call the head of the language, distributed with
ranks between 1 and $k_-$; a second region, identified as the body of the
language, \textcolor{black}{lies} between $k_-$ and $k_+$; and finally the tail, beyond $k_+$.
\textcolor{black}{From the values reported in  Fig. \ref{fig:diversity}, we see that
$9< k_- <22$, while $k_+$ lies between $1832$ and $3099$.}
As
shown in Fig.~\ref{fig:mu:t}, these regions are robust to changes in the historical period considered and to the
data set size (larger for recent years).

 % }}}

The bodies of languages consist of words that have limited change in time.
\textcolor{black}{Based on the size of basic vocabularies, it can be argued that the }``core'' of English is
\textcolor{black}{between 1500 and 3000 words, as mentioned in the 
introduction}, which is consistent with our results. If
we agree that the rank diversity identifies the core (head and body) of
English, then it can be argued that the size of the core of the other five
languages studied is similar~\cite{Hernandez1988Hacia-un-modelo}, which is also
supported by the high similarity across languages in Fig~\ref{fig:diversity}.

The tails of languages are formed by words which vary their rank considerably
in time. This implies that they are more dependent on the text and its domain
than words from the core. It can be assumed that words belonging to
the head and body of languages have a high probability of being used in any
text, while words from the tail would appear only in specific texts and
domains. 

Note that we obtain language cores slightly larger than those proposed by
linguists. This is to be expected, as \textcolor{black}{the Google Books data set treats words forms inflected for different persons, tenses, genders, numbers, cases, and so forth, as distinct items}, while dictionaries count only \textcolor{black}{stems (presented as citation forms, i.e. the basic form that users are most likely to look up)}. \textcolor{black}{For example, the core for English obtained using rank diversity consists of 2448 words, but within these there are only 1760 different \textcolor{black}{stems} in the year 2008.} Moreover, the studied data set contains several proper \textcolor{black}{names} which are not included in basic vocabulary lists. \textcolor{black}{For English, 55 out of 2448 are proper names in 2008. }

\begin{figure}[H] % {{{ Word walks
\centering
     \subfigure{
          \label{fig:word:walks}
         \figura{\includegraphics[width=.45\textwidth]{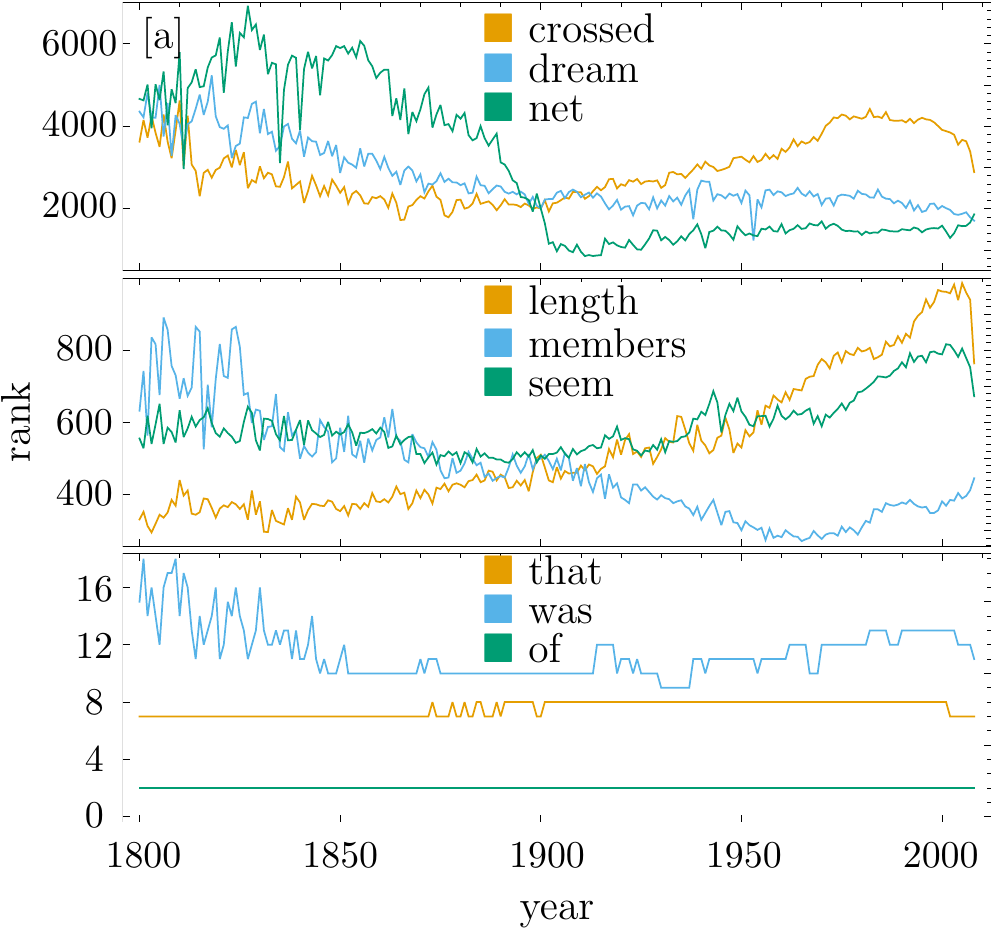}}
}
%     \hspace{.3in}
     \subfigure{
          \label{fig:word:walks:artificial}
          \figura{\includegraphics[width=.45\textwidth]{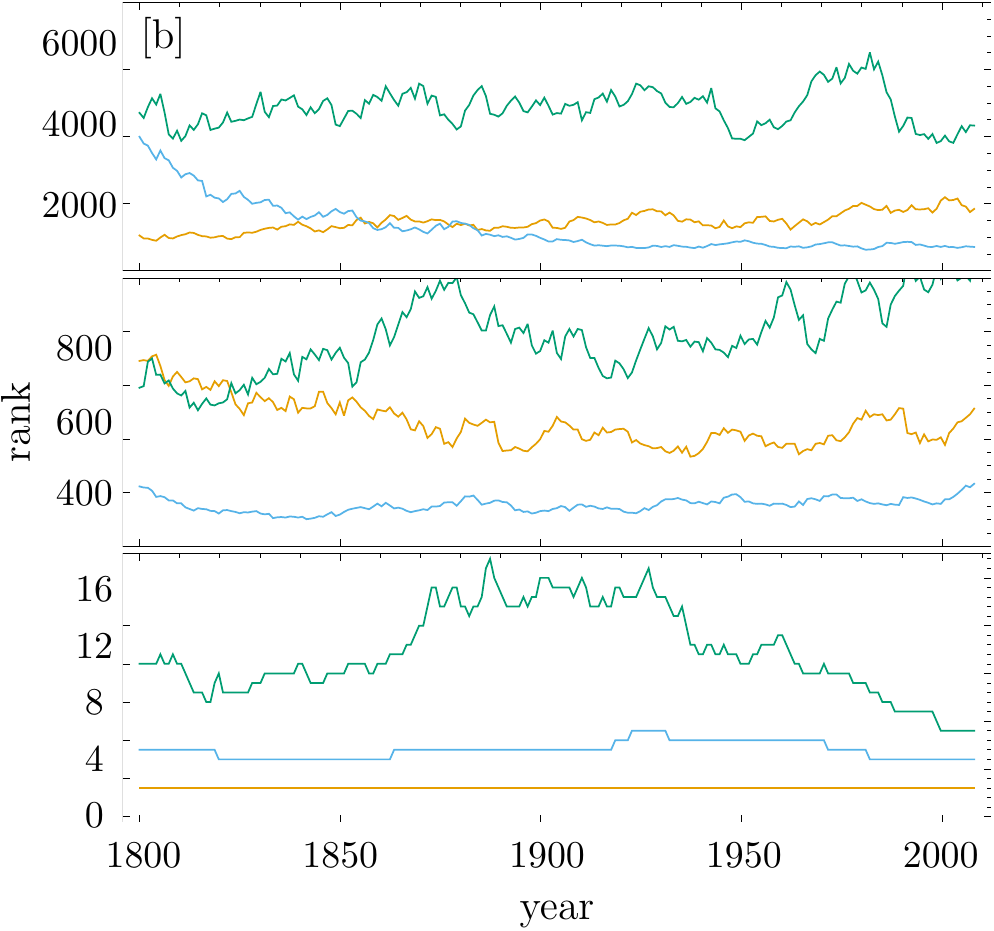}}
}

     \caption{{\bf Rank evolution}. [a]: Evolution of the rank for several
     particular, but random words in different regimes in the English language.
     From bottom to top we show words with initial ranks of order $1$ (head),
     $100$ (body) and $1000$ (tail). [b]: Evolution of the rank for several
     particular, but random words in different regimes, for our scale-free
     Gaussian walker, i.e. the \textcolor{black}{simulated} language we have generated. }
  \label{fig:word:walks:both}
\end{figure}

The rank evolution of particular words in time, belonging to the head, body,
and tail of English is shown in Fig.~\ref{fig:word:walks}. This ratifies the
results shown in Fig.~\ref{fig:diversity}, where \textcolor{black}{low-ranked} words \textcolor{black}{exhibit little variation in time and this variation increases with the rank}. More trajectories are
presented in the SI.  As mentioned above, words from the head \textcolor{black}{vary little over time}. However, the way in which words from the body or tail vary \textcolor{black}{their rank in time}
appears to be similar, although at a different scale. This similarity leads us
to propose a model of rank diversity where the \textcolor{black}{amount of rank variation depends}
only on the rank.

 % }}}

% }}}
\section*{A random walk model for rank diversity} % {{{
\label{sec:model}

\begin{figure}[H] % {{{ Distribution of steps
     \centerline{\figura{\includegraphics[width=.7\textwidth]{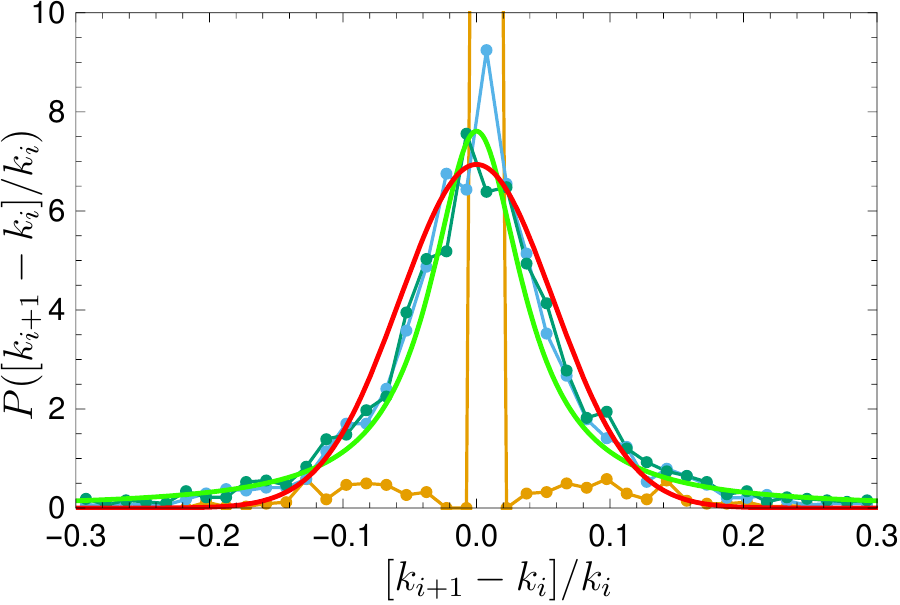}} }
     \caption{  \textcolor{black}{  {\bf Distribution of relative size of 
     frequency changes} $[k_{t+1}-k_t]/k_t$
     in the case of English for words in 
     the head  {\color{colora} $\blacksquare$} (that start with rank between 1 and 10),
     the body  {\color{colorb} $\blacksquare$} (rank between 200 and 210),
     and the tail {\color{colorc} $\blacksquare$} (rank between 5000 and 5010).
%      between 1000 and 1010; 
%      {\color{colorc} $\blacksquare$}, 
%      for words between 5000 and 5010 and 
%      {\color{colord} $\blacksquare$}, 
% \label{eq:walk}
%      for words between 9980 and 9990. 
     Notice that for words in the head, the granularity of the model (\eref{eq:walk}) shows up as large deviations from 
     the Gaussian. 
     For the body and tail, the relative jumps are \textcolor{black}{similar independently} of the
     initial rank of the word.  We also show, as a thick green curve, the
     Lorentzian distribution which best fits the average of the curves for the body and tail. A
     Gaussian, with zero mean and the most common standard deviation $\hat\sigma=0.0575$, is also
     shown in red for comparison (see text for details). The corresponding plot for other languages
     is shown in the supplementary information.}}
  \label{fig:flight:distribution}
\end{figure}

\textcolor{black}{
We consider the relative size of frequency changes, or flights as they are
sometimes called in statistical physics, defined as 
$\left(k_{t+1}-k_{t}\right)/k_{t}$ where $k_t$ is the rank at discrete time
$t$ of a given element. We
present  in Fig.~\ref{fig:flight:distribution} the distribution of these frequency changes
for English, our largest data set, and in Fig.~\ref{fig:allFlightDistributions} in File S1 for all languages.
Notice that, on average, the relative jumps seem to be largely independent of the value
of the rank. 
We propose, based on this fact, a simple model to understand the evolution of rank
diversity of words.}

 % }}}

We shall call this model  a scale-invariant random Gaussian walk,
since a word with rank $k_t$, is
converted to rank $k_{t+1}$ according to the following procedure: One defines
an auxiliary variable $s_{t+1}$ at time $t+1$ by the relation
\begin{equation}
s_{t+1}=k_t+G\left(0, k_t \hat\sigma\right),
\label{eq:walk}
\end{equation}
where $G(0, \tilde\sigma)$ is a Gaussian random number generator of width
$\tilde\sigma$ and mean $0$. This means that the random variable
$s_{t+1}$ has a width distribution proportional to $k_{t}$. Words with very low
ranks will change very slowly or not at all, while those with higher $k$ have
a larger rank variation in time, as reflected by $d(k)$. 
Once the values of $s_{t+1}$ for all words are obtained, they are ordered
according to their magnitude. This new order gives new rankings, \emph{i.e.}
the $k$ values at time $t+1$. 
\textcolor{black}{
There is a small correlation of the jumps between different times in this
model. This is consistent with the observed behavior of the six languages
dealt with here, as can be seen in Fig.~\ref{fig:word:walks:correlation} in File S1.
The only parameter in the model is the width
$\hat \sigma$, which is the most common standard deviation of the relative frequency changes of each data set.}

A word of caution must be said. In Fig.~\ref{fig:flight:distribution}, two
curves are plotted. In green, a Lorentzian distribution, and in red a Gaussian
distribution, both centered at zero, and with a width obtained by best fit to
the data presented here.  Although the Lorentzian fits these data somewhat
better than the Gaussian, we use the latter in our model, since  the long tails
of the Lorentzian would yield long flights in words (not observed in the
historical data) and a very different function $d(k)$. One should
recall that the Lorentzian does not have a finite second moment, so this might
be the reason for this distribution to be inadequate. It is probable that a
truncated Lorentzian could be a better choice, but we leave this detail open as
a possible refinement to our model.

\begin{figure}[H] % {{{ Diversidad simulada y las otras normalizadas
     \centering
%           \label{fig:d:synthetic}
         \figura{\includegraphics[width=.7\textwidth]{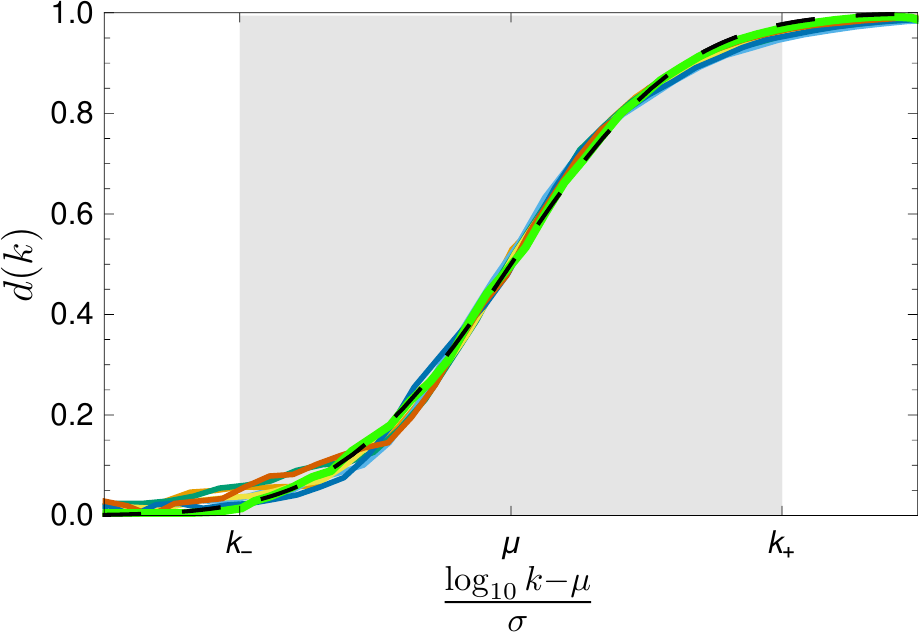}}
     \caption{\textcolor{black}{{\bf Rank diversity for the simulated language.}  The green curve
     represents the diversity corresponding to the language dynamics of a
     single realization of the Gaussian random walk model. We also include data for all
     languages studied, but normalized so that $k_{\pm}$ coincide. The ansatz for
     the rank diversity is plotted as a parameter-free cumulative of a Gaussian with 
     zero mean and unit variance as a dashed black curve. }
     }
  \label{fig:diversity:normalized}
\end{figure}

\textcolor{black}{With this model we have produced the evolution of a random simulated
language; see~\cite{steels1997synthetic} for other approaches.
Fig.~\ref{fig:word:walks:artificial} shows examples of rank trajectories at
different scales, exhibiting similarities with those of actual words shown in
Fig.~\ref{fig:word:walks}.  Moreover, if its diversity $d(k)$ is
calculated with the $\hat \sigma$ corresponding to the most popular width of
the distribution of relative size of flights for all words in the English
language from 1800 to 2008, the results coincide with the sigmoid obtained for
all six languages analyzed, as shown in Fig.~\ref{fig:diversity:normalized}. 
}

 % }}}

% }}}
\section*{Discussion} % {{{

\textcolor{black}{Within statistical linguistics, the frequency-rank distributions of several languages of European origin have been analyzed for many years now. However, no simple model
can reproduce the detailed properties of this distribution (see SI). In particular,
there has been the proposal that there exist two different regimes
for ranks, \textcolor{black}{but these regimes have not been satisfactorily validated in the empirical data}.
Due to these difficulties we have been \textcolor{black}{led} to introduce a statistical measure,
which we have called rank diversity, to describe the statistical properties
of natural languages. A simulated random language was generated
which reproduces \textcolor{black}{the observed features quite well}. 
}

Our random walk model mimics the evolution of languages to produce \textcolor{black}{a simulated}
rank diversity which closely matches that of historical data. We consider that
statistical similarities across languages and the simplicity of the model to
reproduce them sufficient evidence to claim that rank diversity \textcolor{black}{of} words \textcolor{black}{is} universal. \textcolor{black}{This does not imply that all languages have the same rank diversity curves, but that the rank diversity distribution of all the languages studied here can be fitted properly with \eref{eq:sigmoid}. Certainly, different languages have different curves that fit them better, just as different exponents fit better a Zipf distribution of different languages. For the languages studied, $1.6\leq\mu\leq2.1$ and $0.4\leq\sigma\leq0.6$}.

This universality \textcolor{black}{could} be used to favor nativist explanations of human
language~\cite{chomsky1965aspects,hauser2002}, where language \textcolor{black}{is claimed to be }determined by
innate constraints. However, the \textcolor{black}{high-ranked} diversity of language tails could be
used in favor of adaptationist explanations \textcolor{black}{as well}~\cite{BBS:7242672}, as the precise
rank of tail words is highly contingent. In recent years, explanations of human
language relating biological evolution (genetically encoded innate properties)
and learning (epigenetical adaptation) with culture have gained
strength~\cite{%deacon1998symbolic,
kirby1999function,Kirby20032007,%Kirby30072008,
Chater27012009}.
Even so, few assumptions are necessary to explain some general aspects of the evolution of human
languages~\cite{Nowak06071999}. The present work shows that the evolution of
word \textcolor{black}{frequency} can be explained with Gaussian random walks, where the size of the
change in word \textcolor{black}{frequency} is proportional to its rank, \emph{i.e.} frequent words
change less than infrequent words. This explanation does not require innate
properties, adaptive advantages, nor culture. \textcolor{black}{This does not imply that the latter are irrelevant for other aspects of language evolution. Note that our study is \textcolor{black}{carried out} at a statistical level. We do not
address \textcolor{black}{syntactic, semantic,} and grammatical aspects of human
language~\cite{steels1995self, Sandler15022005, Gell-Mann18102011,
citeulike:12174877}\textcolor{black}{, which are certainly important}.
}

Why does the rank diversity approach a lognormal distribution? Which processes
and mechanisms are required for this?  There is one condition for a variable to
have a lognormal distribution. \textcolor{black}{This condition is that the variable should be the result of a
high number of different and independent causes which produce
positive effects composed multiplicatively. Thus, each cause has a negligible effect on the
global result~\cite{Brockmann13122013}.} Our Gaussian random walk model supports this as a suitable
explanation: the statistical distribution of $d$ is always lognormal, there is
a high number of components (words), each word has a negligible effect compared
to the language properties, \emph{i.e.} large changes in word \textcolor{black}{frequency} (ranking)
do not cause large changes in \textcolor{black}{the statistical properties of each language}, and the rank of each word
is partially a cumulative product of its rank in previous times, as expressed
in \eref{eq:walk}.  Languages \textcolor{black}{statistically comply with} these dynamics, and that
serves as an explanation for their evolution and structure.

\begin{figure}[H] % {{{ Chess
\centering
        \figura{\includegraphics[width=.7\textwidth]{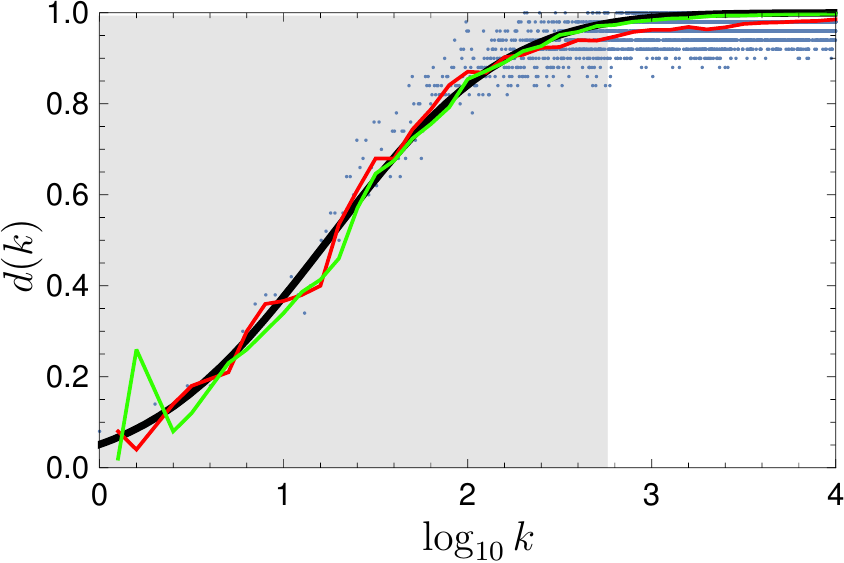}}
     \caption{\textcolor{black}{{\bf Rank diversity of male chess players} obtained from
     the trimestral FIDE rankings from April, 2001 to May, 2012 ($\Delta
     t=50$), considering the first 10,000 ranks. Blue dots show rank diversity,
     windowed in the red line. The black line shows the sigmoid fit with
     $\mu=1.24$ and $\sigma=0.76$. The green line shows a simulation with
     $\hat\sigma=0.18$. Notice that there is no head as $\mu-2\sigma <0$. This is to be expected, as many players enter and leave the ranking during the years considered.}}
     \label{fig:chess}
\end{figure}

\textcolor{black}{In} future work, it will be relevant to study the rank diversity of $n$-grams
with $n>1$~\cite{Ha02extensionof}, other linguistic corpora and phenomena with
dynamic rank distributions~\cite{blumm2012dynamics, batty2006rank, CPLX:CPLX20156,
hausmann2014atlas} and more generally with temporal
networks~\cite{gross2009adaptiveNets, Gautreau02062009,
Perra2012Activity-driven, Holme2012TemporalNets}. 
\textcolor{black}{
A specific example \textcolor{black}{would be} the ranking of chess players, given by the World Chess Federation (F\'ed\'eration Internationale des \'Echecs). \textcolor{black}{The rank diversity in this case is provided} in figure \ref{fig:chess}, \textcolor{black}{which shows that the sigmoid}
is appropriate also for this case. 
}

 % }}}

% }}}
% \section*{Acknowledgments} % {{{
% 
% \textcolor{black}{We are grateful to the editor and the two anonymous referees for their useful comments.
% %G.C. received support from project UNAM-PAPIIT IN107414. C.G. was supported by SNI membership 47907 of CONACyT, Mexico. C.P. received support from the projects CONACyT 153190 and  UNAM-PAPIIT IA101713 and IN111015.
% }
%  % }}}
%\section*{References}

% }}}
\clearpage
\newcommand{\mcN}{\mathcal{N}}
\renewcommand{\rme}{\mathrm{e}}
\renewcommand{\rme}{\mathrm{e}}
\renewcommand{\rmd}{\mathrm{d}}
\renewcommand{\eref}[1]{equation~(\ref{#1})} \renewcommand{\sref}[1]{section~\ref{#1}}
\renewcommand{\fref}[1]{figure~\ref{#1}}     \renewcommand{\tref}[1]{table~\ref{#1}}
\newcommand{\Eref}[1]{Equation~(\ref{#1})} \newcommand{\Sref}[1]{Section~\ref{#1}}
\renewcommand{\Fref}[1]{Figure~\ref{#1}}     \renewcommand{\Tref}[1]{Table~\ref{#1}}

\newcommand{\headernountable}{  & English &  German & French & Italian & Spanish & Russian \\ \hline}
\newcommand{\headernountabletwo}{ \begin{tabular}{|  c | c|  c|  c|  c|  c|  c|  c|  c|  c|  c|  }  \hline}

\renewcommand{\thesection}{S\arabic{section}}\renewcommand{\theequation}{S\arabic{equation}} % }}}
\section*{Supporting Information} % {{{
\section{Models for rank-frequency distributions} %{{{

% As described by Zipf~\cite{zipf}, t
The rank-frequency distributions of words for
different languages are very similar to each other, as shown in
Fig.~\ref{fig:several:fixes:year}. The distributions are also similar across
centuries, as shown in Fig.~\ref{fig:english:example}.

\begin{figure*} % {{{ 2000, several languages

\subfigure{
          \label{fig:several:fixes:year}
          \includegraphics[width=.45\textwidth]{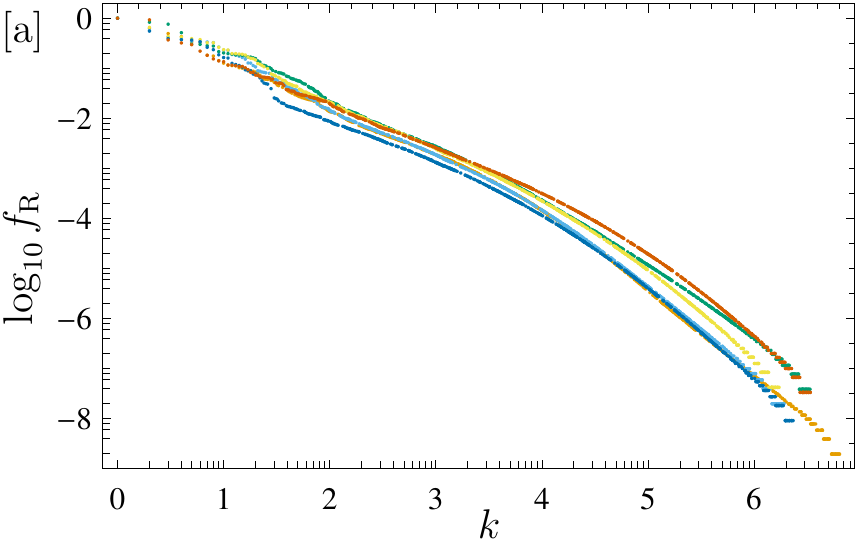}}
    \hspace{.3in}
     \subfigure{
          \label{fig:english:example}
          \includegraphics[width=.45\textwidth]{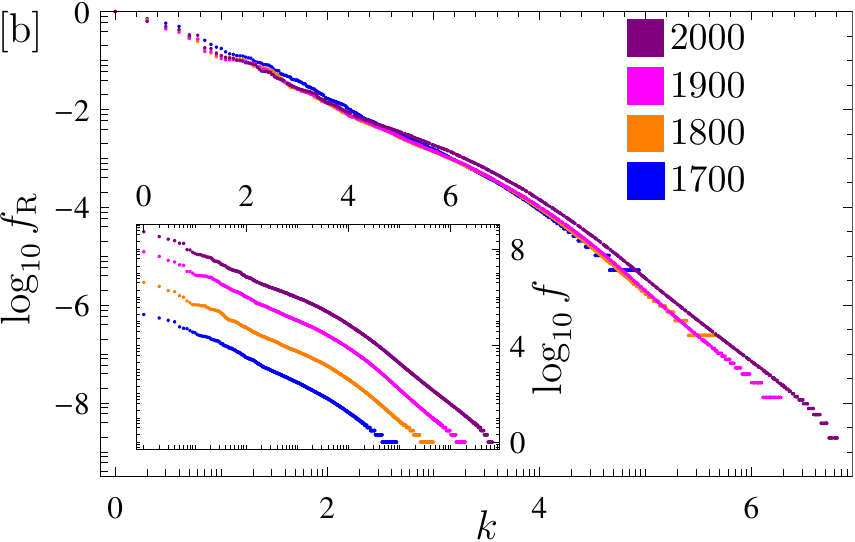}}
\caption{Rank distributions of words according to frequency. [a]: Normalized
word frequency $f_\textup{R}$ as a function of the rank $k$ for several
languages for books published in the year 2000.
     The color code for languages is as follows: 
     {\color{colorb} $\blacksquare$} for French, 
     {\color{colorc} $\blacksquare$} for German, 
     {\color{colord} $\blacksquare$} for Italian, 
     {\color{colora} $\blacksquare$} for English,
     {\color{colore} $\blacksquare$} for Spanish, and
     {\color{colorf} $\blacksquare$} for Russian.
      [b]: Word frequency $f_\textup{R}$ as a function of
the rank $k$ for English and several years, normalized so that the most
frequent element has relative frequency one. In the inset, the unnormalized
frequency $f$ is shown.} \label{fig:Zipfians}
\end{figure*} % }}}

We present five different distributions with distinct origins,
though all of them containing the common factor $\frac{1}{k^a}$.
The distributions are:
% 
% We present five different distributions with distinct origins, though all of
% them containing the common factor $\frac{1}{k^a}$.  The distributions are: 
\begin{align}
m_1(k) &=\mcN_1  \frac{1}{k^a}, \label{eq:m1}\\
m_2(k) &=\mcN_2  \frac{\rme^{-b(k-1)}}{k^a},  \label{eq:m2}\\
m_3(k) &=\mcN_3  \frac{(\bar N +1 -k)^\alpha}{k^a},  \label{eq:m3} \\
m_4(k) &=\mcN_4 \frac{(\bar N +1 -k)^\alpha\rme^{-b(k-1)}}{k^a},  \label{eq:m4}\\
m_5(k) &=\mcN_5 \begin{cases}
\frac{1}{k} & k \le k_c  \\
\frac{k_c^{a-1}}{k^{a}} & k > k_c  
\end{cases}
 \label{eq:m5}
\end{align}
where $\mcN_i$ are normalization factors, depending on the parameters $a$, $b$,
and $\alpha$  of the different models, and $\bar N$ is the total number of words. 

In Fig.~\ref{fig:fits} we compare the fit of these distributions with the
observed curves. 
It can be seen that none of the distributions reproduces closely the dataset. We
calculated for all fits the $\chi^2$ test with similar results. The best 
value corresponds to the fit proposed in~\cite{PhysRevX.3.021006}, namely the
double Zipf model (\eref{eq:m5}).
In all cases we studied the $p$-value of the 
data, needed for an appropriate interpretation of the goodness of the fit. In
all cases, that is for all years, all languages and all models, this number was smaller
than machine precision. 
This shows that none of these models captures satisfactorily the data
behavior.
 
\begin{figure*} % {{{ Fit
\includegraphics[width=\textwidth]{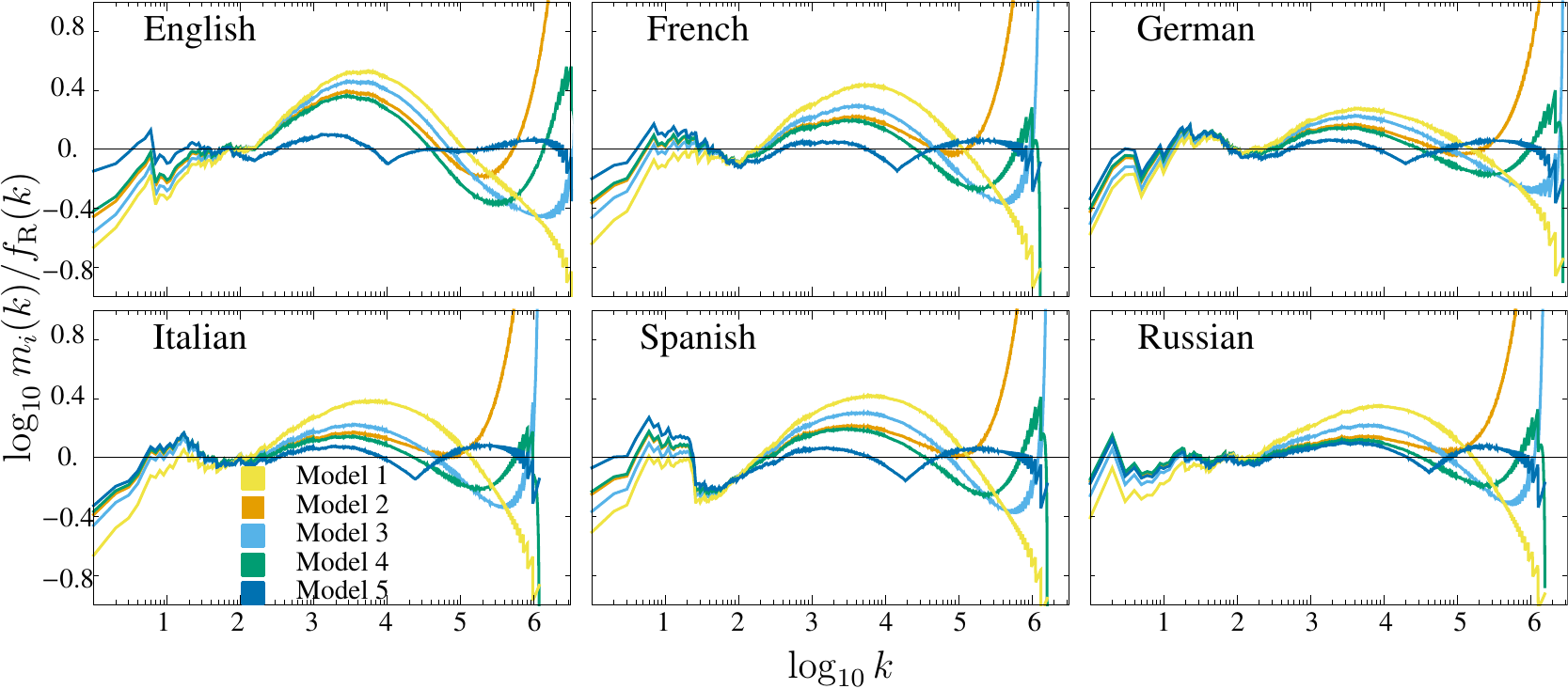}
\caption{ Comparison between the different models,
\eref{eq:m1}--\eref{eq:m5}, and the frequency of rank distribution. 
We use the data for the year 2000 and all languages under
consideration.  The logarithm base 10 of the ratio of the observed
values and the model is plotted.  It can be appreciate that different models fit better in
different regions. However there is no model that fits all languages and all
regions much better than the others. 
}
\label{fig:fits}
\end{figure*} % }}}

The origin of some of these models is similar. The following discussion shows
how they can be encompassed in a common formulation.

Given a set of words forming a text, one can evaluate
the number of times $N(k,t)$ that a certain word appears
with the rank $k$ at time $t. $ If $B(k)$and $D(k)$
denote, respectively, the probability per unit time that a word enters
or leaves the rank $k$, we have: 
% 
% Stochastic dynamics is a natural framework for the study of the evolution of
% complex networks~\cite{RevModPhys.74.47, Jensen08082008, PhysRevE.62.8466}.
% Given a network with nodes labeled by their connectivity, changes in their
% distribution can be studied in terms of death-birth one-step stochastic
% processes. More specifically, if $B(k)$ and $D(k)$
% denote, respectively, the probabilities per unit time that a node with $k$
% edges enters or leaves the network at time $t,$ the time evolution of
% the number of nodes with $k$ edges,
% $N(k,t)$, may be written in the general form
\begin{multline}
\frac{\partial}{\partial t}N(k,t)=
\left\{ \xi(k)-F[\Sigma(t)]\right\} 
   N(k,t) \\
+\Big\{ 
  D(k+1)N(k+1,t) 
  +B(k-1)N(k-1,t)
  -\left[D(k)+B(k)\right]N(k,t)\Big\}. 
\label{eq:1}
\end{multline}
Here the two terms on the r.h.s. within the first curly
brackets describe, respectively, the local growth rate and the overall
decrease rate acting on $N(k,t).$ The total number
of words at a given time $t$ is 
$\Sigma(t)$
and $F$ is
a function that determines global constraint features that refer to the total
number of words. The terms within
the second curly brackets, indicate the balance arising from the birth $B(k)$
and death $D(k)$ contributions of first neighbor words with $k\pm1$ ranks
at time $t$. If we consider the total number of words at a given
time $t$ to be a fixed quantity, 
\begin{equation}
\Sigma(t)=\sum_{k}N(k,t)
\label{eq:2}
\end{equation}
we can define the probability density of finding a word with
rank $k$, or relative frequency distribution, by
\begin{equation}
n(k,t)\equiv\frac{N(k,t)}{\Sigma(t)}.
\label{eq:3}
\end{equation}
Substitution of \eref{eq:2} and \eref{eq:3} into \eref{eq:1} leads to
\begin{equation}
\frac{d}{dt}\Sigma(t)=\langle\xi(k)\rangle-F[\Sigma(t)]\Sigma(t),
\label{eq:4}
\end{equation}
where the bracket indicates a sum over all $k$ weighted by $n(k,t).$
We assume, for simplicity, that $\langle\xi(k)\rangle$ is a linear
function of the number of edges $k$, so that $\langle\xi(k)\rangle$
= $\xi_{0}+\xi_{1}\langle k\rangle,$ where $\xi_{0}$ and $\xi_{1}$ are constants.
Then \eref{eq:1} reduces to the following master equation for a one step
process: 
\begin{multline}
\dot{n}(k,t)=V(k)n(k,t)\\
+\left\{ D(k+1)n(k+1,t)+B(k-1)n(k-1,t)-\left[D(k)+B(k)\right]n(k,t)\right\}, 
\label{eq:5}
\end{multline}
where $\dot{n}(k,t)\equiv\partial n(k,t)/\partial t$
and the effective potential
$V(k)\equiv\xi(k)-\langle \xi(k)\rangle$
has the property
\begin{equation}
\langle V(k)\rangle=0.
\label{eq:6}
\end{equation}
In what follows we shall only consider the case $\xi(k)=\xi_{0},$
so \eref{eq:5} reduces to the general form of the master equation for
a one step process, 
\begin{equation}
\dot{n}\left(k,t\right)=D\left(k+1\right)n\left(k+1,t\right)+
B\left(k-1\right)n\left(k-1,t\right)-
 \left[D(k)+B(k)\right]n(k,t).
\end{equation}
If the changes in $k$ are small and we are only interested
in solutions $n\left(k,t\right)$ that vary slowly with $k$, then
$k$ may be treated as a continuous variable and we obtain the Fokker-Planck
equation:
\begin{equation}
\frac{\partial n\left(k,t\right)}{\partial t}=-\frac{\partial}{\partial k}\left[g\left(k\right)n\left(k,t\right)\right]+\frac{1}{2}\frac{\partial^{2}}{\partial k^{2}}\left[f\left(k\right)n\left(k,t\right)\right],
\end{equation}
where $f(k)=B(k)+D(k)$ and $g(k)=B(k)-D(k)$.  For the stationary solutions
$m(k)$, we have the equation
\begin{equation}
g(k)m(k)=\frac{1}{2}\frac{d}{dk}\left[f(k)m(k)\right].
\end{equation}
If we approximate $g(k)/f(k)$ by Pad\'e approximants $g_n(k)/f_n(k) =A_0+\sum_{k=1}^n\frac{A_k}{(k+c_k)}$,
the stationary solution becomes
\begin{equation}
m(k)=\mathcal{N}\textrm{exp}\left(A_{0}k\right)\prod_{k=1}^{n}(k+c_{k})^{-A_k}.
\end{equation}
If we assume the simplest expression for $D(k)$ and
$B\left(k\right)$ transition probabilities
\begin{align}
D(k)&=\lambda_1(c_1+k)(N_1-k),\\
B(k)&=\lambda_2(c_2+k)(N_2-k)
\end{align}
then
\begin{equation}
m(k)=\mathcal{N}\textrm{exp}(A_{0}k)\frac{\left(\bar{N}-k\right)^{b}}{(\bar{c}+k)^a},
\end{equation}
where
$\bar{N}=\frac{1}{2}\left(N_{1}+N_{2}\right)$,
$\bar{c}=\frac{1}{2}\left(c_1+c_2\right)=1$,
$a=c_1-c_2+1$, and
$b= N_{1}-N_{2}-1$. Also we must remember that in our case $k$ starts at one.
Then if $A_{0}=b=0,$ we have the Zipf model;
% \begin{equation}
% m_{1}\left(k\right)=\mathcal{N}/k^{a}.
% \end{equation}
when $A_{0}\neq0$ and $b=0,$ the $\gamma$ model is gotten  (\eref{eq:m2});
% \begin{equation}
% m_{2}\left(k\right)=\mathcal{N}\frac{\textrm{exp}\left(-dk\right)}{k^{a}}.
% \end{equation}
if $A_{0}=0$ but $b\neq0$ the $\beta$ model is obtained  (\eref{eq:m3});
% \begin{equation}
% m_{3}\left(k\right)=\mathcal{N}\frac{\left(\bar{N}-k\right)^{b}}{k^{a}}.
% \end{equation}
finally, if $A_{0}$ and $b$ are different from 0, we have the general
$\beta\gamma$ model  (\eref{eq:m4}). 
% \begin{equation}
% m_{4}\left(k\right)=N\frac{exp\left(-dk\right)\left(\bar{N}-k\right)^{b}}{k^{a}}.
% \end{equation}

These and additional results could be obtained using the complex network
language~\cite{RevModPhys.74.47, Jensen08082008, PhysRevE.62.8466}. 

With respect to the distribution of \eref{eq:m5}, the derivation
given in \cite{PhysRevX.3.021006} is based on the following assumptions.
The existence of two word regimes: A language core containing words
with low rank and do not affect the birth of new words, and the remaining 
high ranked words which reduce the probability of new words to be used.

% }}}
\section{Variation of words in time} % {{{

Table S1 shows the most frequent words for the year 2000 with their
translation and relative frequency. Notice that these are very similar across languages.  Table
S2 shows the most frequent nouns for the years 1700, 1800, 1900, and 2000.
There are  similarities across languages and across centuries, but also
important differences.

\begin{landscape}
\begin{table} % {{{ Most often words
\caption{Lowest--rank words for several languages in books published during the year 2000, together with their translation to English and their relative frequency.}
\begin{tabular}{|  c|  | c|  c|  c|  c|  c|  c|  c|    c|  c|  }  \hline
rank & English & German & French &  Italian & Spanish & Russian \\\hline \hline
1  &  the, 0.065530	&	der, the, 0.038512	&	de, of, 0.057225	&	di, of, 0.041518	&	de, of, 0.073063	&	и, and, 0.053961 \\\hline
2  &  of, 0.036769	&	die, the, 0.036010	&	la, the, 0.035222	&	e, and, 0.028107	&	la, the, 0.043297	&	в, in, 0.053922 \\\hline
3  &  and, 0.029289	&	und, and, 0.028087	&	et, and, 0.024466	&	la, the, 0.020308	&	en, in, 0.029059	&	на, on, 0.020190 \\\hline
4  &  to, 0.025264	&	in, in, 0.020607	&	le, the, 0.022384	&	che, that, 0.017861	&	y, and, 0.028908	&	не, not, 0.017334 \\\hline
5  &  in, 0.021769	&	von, of, 0.011277	&	les, the, 0.021076	&	il, the, 0.017702	&	el, the, 0.027771	&	что, what, 0.011770 \\\hline
6  &  a, 0.020715	&	den, the, 0.011012	&	à, to, 0.019951		&	in, in, 0.017357	&	que, that, 0.026713	&	по, by, 0.010202 \\\hline
7  &  is, 0.010712	&	zu, to, 0.010488	&	des, of, 0.019212	&	a, to, 0.014067		&	a, to, 0.019706		&	к, to, 0.008559 \\\hline
8  &  that, 0.010529	&	des, of, 0.010102	&	en, in, 0.014334	&	del, of, 0.013403	&	los, the, 0.018039	&	как, as, 0.008027 \\\hline
9  &  for, 0.008975	&	das, the, 0.009806	&	du, of, 0.012991	&	della, of, 0.010876	&	del, of, 0.013492	&	а, and, 0.007745 \\\hline
10  &  as, 0.007396	&	im, in the, 0.007418	&	un, a, 0.011112		&	per, for, 0.010480	&	se, oneself, 0.012448	&	о, about, 0.006824 \\\hline
11  &  it, 0.006832	&	mit, with, 0.007403	&	une, a, 0.010825	&	un, a, 0.009949		&	las, the, 0.012294	&	из, of, 0.006356 \\\hline
12  &  with, 0.006707	&	sich, itself, 0.007337	&	dans, in, 0.010145	&	non, not, 0.008645	&	por, by, 0.009908	&	его, his, 0.005911 \\\hline
13  &  was, 0.006576	&	ist, is, 0.007197	&	que, that, 0.009896	&	si, oneself, 0.008515	&	un, a, 0.008824		&	для, for, 0.005822 \\\hline
14  &  on, 0.006289	&	auf, on, 0.007047	&	qui, who, 0.008609	&	è, is, 0.008501		&	con, with, 0.008469	&	от, from, 0.005769 \\\hline
15  &  not, 0.005970	&	nicht, not, 0.006875	&	par, by, 0.007494	&	una, a, 0.007891	&	una, a, 0.007863	&	он, he, 0.005538 \\\hline
16  &  be, 0.005671	&	für, for, 0.006874	&	est, is, 0.007258	&	le, the, 0.007852	&	no, no, 0.007547	&	но, but, 0.005324 \\\hline
17  &  by, 0.005440	&	eine, a, 0.006757	&	pour, for, 0.007027	&	i, the, 0.007626	&	para, for, 0.006877	&	я, I, 0.005097 \\\hline
18  &  i, 0.005212	&	als, as, 0.006521	&	il, it, 0.006749	&	con, with, 0.006734	&	su, its, 0.006597	&	это, this, 0.004925 \\\hline
19  &  are, 0.004928	&	dem, the, 0.005723	&	au, to the, 0.006429	&	da, from, 0.006258	&	es, is, 0.006086	&	за, for, 0.004623 \\\hline
20  &  this, 0.004916	&	auch, also, 0.005630	&	a, has, 0.005504	&	nel, in, 0.005184	&	al, to the, 0.005855	&	у, at, 0.003862 \\\hline

\end{tabular}\hfill
\label{table:most:often:words}
\end{table} % }}}
\end{landscape}

\begin{table} % {{{ Evolution of nouns
\caption{Lowest ranked \textbf{nouns} for different years (top left cell) and different languages. Note that some words are used not only as nouns, which can give them a higher rank. For example, \emph{\'et\'e} in French is summer, but also the past participle of \emph{\^etre} (to be).}
\scriptsize
\headernountabletwo 1700 \headernountable{} 1	&	god	&	Erfahrung, experience 	&	fait, fact 	&	rei, king 	&	fe, faith 	&	день, day 	\\\hline
2	&	man	&	Gottesfurcht, fear of god 	&	dieu, god 	&	sez, section 	&	señor, mr. 	&	города, city 	\\\hline
3	&	men	&	Derselben, the same 	&	point, point 	&	civ, civil code 	&	cardenal, cardinal 	&	капитанъ, captain 	\\\hline
4	&	people	&	Denselben, the same 	&	corps, body 	&	giudice, judge 	&	rey, king 	&	года, year 	\\\hline
5	&	first	&	Dieselbe, the same 	&	amour, love 	&	parte, part 	&	dios, god 	&	утру, morning 	\\\hline
6	&	things	&	Dieselben, the same 	&	car, car 	&	comma, paragraph 	&	solo, single 	&	полки, shelves 	\\\hline
7	&	time	&	Denselben, the same 	&	Reims, Reims 	&	lavoro, work 	&	tiempo, time 	&	ночь, night 	\\\hline
8	&	world	&	Menschen, people 	&	 temps, time	&	diritto, right 	&	san, saint 	&	лошадей, horses 	\\\hline
9	&	thing	&	Alter, age	&	homme, man 	&	art, article 	&	duque, duke 	&	городъ, city 	\\\hline
10	&	power	&	Jugend, youth 	&	roy, king 	&	sentenza, judgment 	&	ácido, acid 	&	вечеру, evening 	\\\hline
   
\hline \hline 1800  \headernountable{} 1	&	time	&	Nichts, nothing 	&	fait, fact	&	era, era 	&	dios, god 	&	время, time 	\\\hline
2	&	king	&	Zeit, time        	&	point, point 	&	parte, part	&	parte, part 	&	году, year 	\\\hline
3	&	man	&	Art, type 	&	été, summer 	&	tempo, time	&	tiempo, time 	&	день, day 	\\\hline
4	&	god	&	Derselben, the same	&	eau, water 	&	prima, first	&	solo, single 	&	года, year 	\\\hline
5	&	first	&	Menschen, people 	&	partie, part 	&	stato, state 	&	señor, mr. 	&	времени, time 	\\\hline
6	&	part	&	Allein, alone 	&	corps, body 	&	città, city 	&	hombre, man 	&	людей, people 	\\\hline
7	&	men	&	Natur,  nature 	&	temps, time 	&	repubblica, republic 	&	cuerpo, body 	&	города, city 	\\\hline
8	&	general	&	--- 	&	terre, land 	&	cose, things 	&	vida, life 	&	образомъ, way 	\\\hline
9	&	people	&	---	&	nombre, number 	&	fatto, fact 	&	modo, mode 	&	земли, land 	\\\hline
10	&	place	&	---	&	homme, man 	&	luogo, place 	&	hombres, men 	&	будетъ, will 	\\\hline
 
\hline \hline 1900  \headernountable{} 1	&	time	&	Selbst, even 		&	été, summer 	&	era, era	&	señor, mr. 		&	время, time 	\\\hline
2	&	man	&	Jahre, years 		&	fait, fact 	&	parte, part 	&	parte, part 		&	года, year 	\\\hline
3	&	first	&	Weise, wise 		&	point, point 	&	stato, state 	&	ley, law 		&	жизни, life 	\\\hline
4	&	life	&	Ersten, first 		&	temps, time 	&	legge, law 	&	gobierno, government 	&	времени, time 	\\\hline
5	&	men	&	Recht, right 		&	cas, case 	&	prima, first 	&	estado, state 		&	образомъ, way 	\\\hline
6	&	day	&	Art, type 		&	droit, right 	&	fatto, fact 	&	derecho, right 		&	будетъ, will 	\\\hline
7	&	old	&	Einzelnen, individual 	&	loi, law 	&	tempo, time 	&	años, years 		&	томъ, volume 	\\\hline
8	&	years	&	Frage, question		&	partie, part 	&	vita, life 	&	año, year		&	году, year 	\\\hline
9	&	work	&	Nichts, nothing 	&	Paris, Paris 	&	anni, age 	&	ciudad, city 		&	права, right 	\\\hline
10	&	people	&	--- 			&	France, France 	&	Italia, Italy 	&	artículo, article 	&	право, right \\\hline
 
\hline \hline 2000  \headernountable{} 1	&	time	&	Deutschen, German 		&	fait, fact 		&	era, era 	&	parte, part		&	время, time 	\\\hline
2	&	first	&	Jahre, years 			&	été, summer 		&	parte, part 	&	años, years old 	&	том, volume 	\\\hline
3	&	people	&	Menschen, people 		&	paris, Paris 		&	stato, state 	&	estado, state 		&	года, year 	\\\hline
4	&	work	&	Frage, question 		&	temps, time 		&	prima, first 	&	vida, life 		&	федерации, federation 	\\\hline
5	&	way	&	Deutschland, Germany 		&	pays, country 		&	anni, years 	&	años, years 		&	жизни, life	\\\hline
6	&	life	&	Jahren, years 			&	politique, policy 	&	vita, life 	&	nacional, national 	&	лет, years 	\\\hline
7	&	world	&	Berlin, Berlin 			&	vie, life 		&	tempo, time 	&	tiempo, time 		&	человек, man 	\\\hline
8	&	way	&	Ersten, first  			&	france, France 		&	secondo, second	&	social, social 		&	году, year 	\\\hline
9	&	state	&	Entwicklung, development 	&	travail, work 		&	modo, way 	&	forma, form 		&	раз, time 	\\\hline
10	&	years	&	Arbeit, work 			&	monde, world		&	fatto, fact 	&	política, policy 	&	человека, human 	\\\hline
 
\end{tabular}
\label{table:most:often:content:words}
\end{table} % }}}

Figs.~\ref{fig:traject_eng}--\ref{fig:traject_sim} show rank trajectories of
words for the languages studied, including our simulated language. It can be
seen that the behavior is similar for all languages: words with low rank
(heads) almost do not vary in time. Afterwards the variation in rank depends on
the rank itself, approximating a scale-invariant random walk. Notice that there
is a higher variation at all scales before 1850. Further work is required to
measure how much this variation depends on having less data before 1850 and how
much on language properties of the time.

%%% experiments: keep N constant for different years, see whether differences hold. Also, try to find
%%%correlations between N in time and variation in spaghetti figs. 

\begin{figure*} % {{{ Espagueti
\begin{center}
     \includegraphics[width=.9\textwidth]{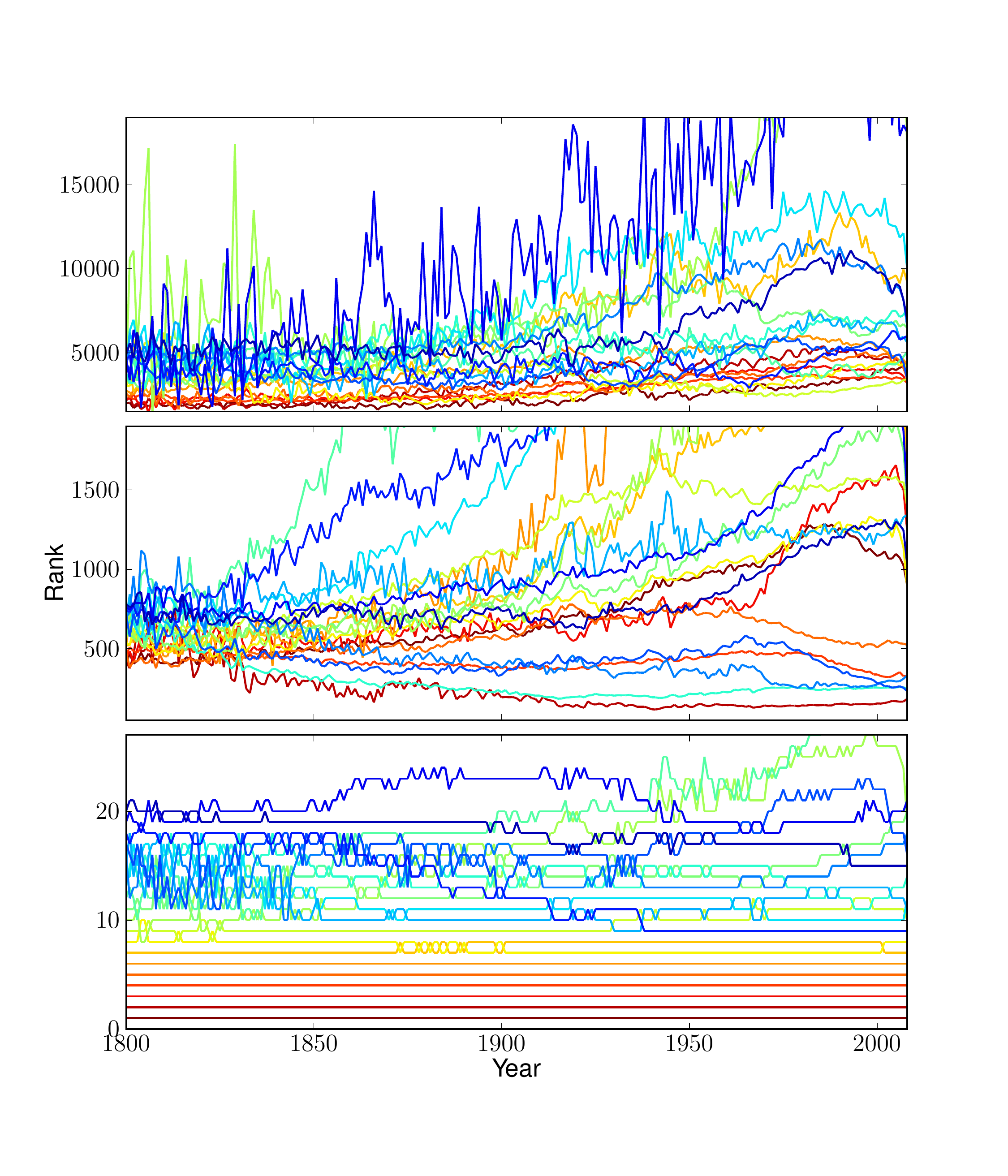}
     \caption{Rank variations in time of twenty words from three different scales for  English. }
  \label{fig:traject_eng}
\end{center}
\end{figure*} % }}}
\begin{figure*} % {{{ Espagueti
\begin{center}
     \includegraphics[width=.9\textwidth]{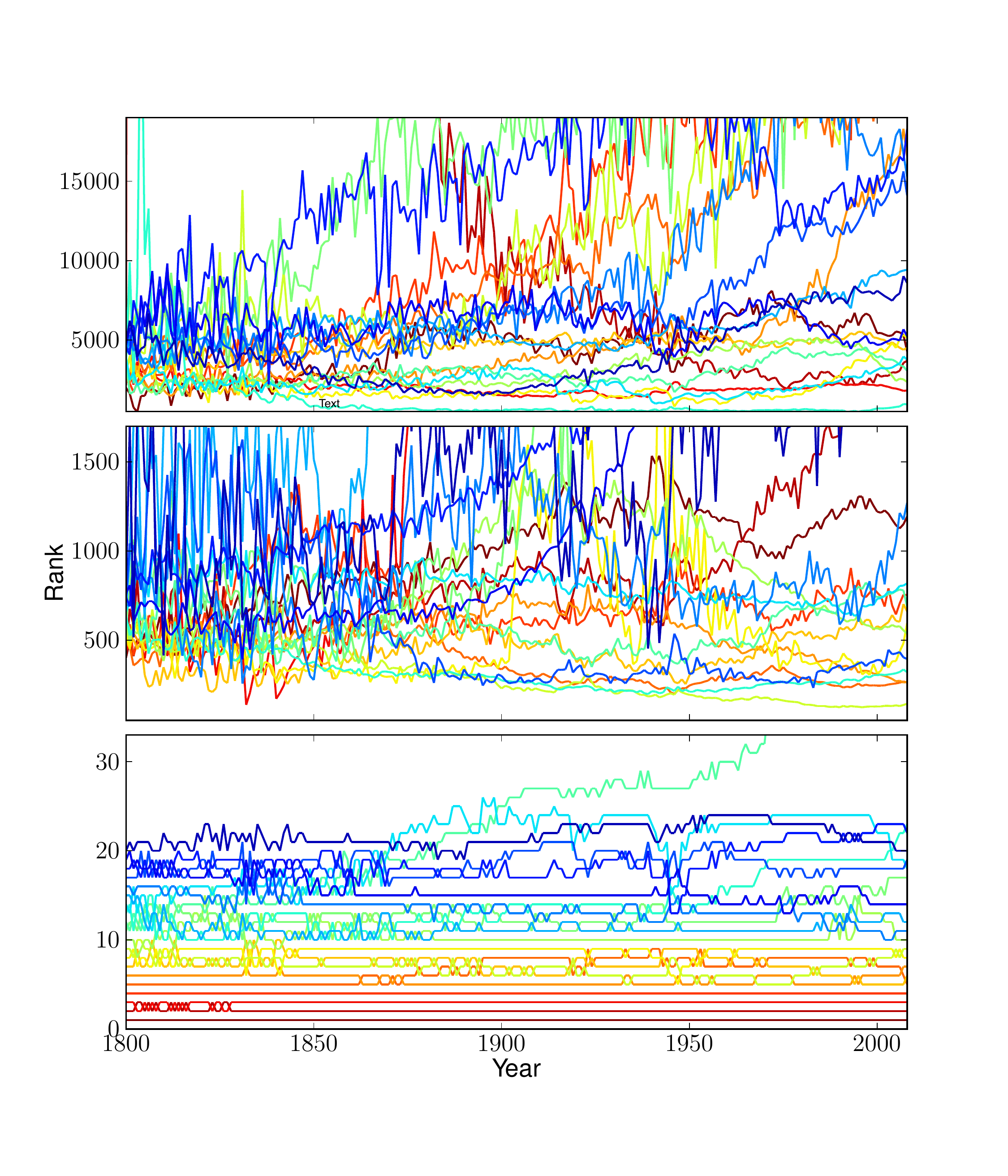}
     \caption{Rank variations in time of twenty words from three different scales for  German. }
  \label{fig:traject_ger}
\end{center}
\end{figure*} % }}}
\begin{figure*} % {{{ Espagueti
\begin{center}
     \includegraphics[width=.9\textwidth]{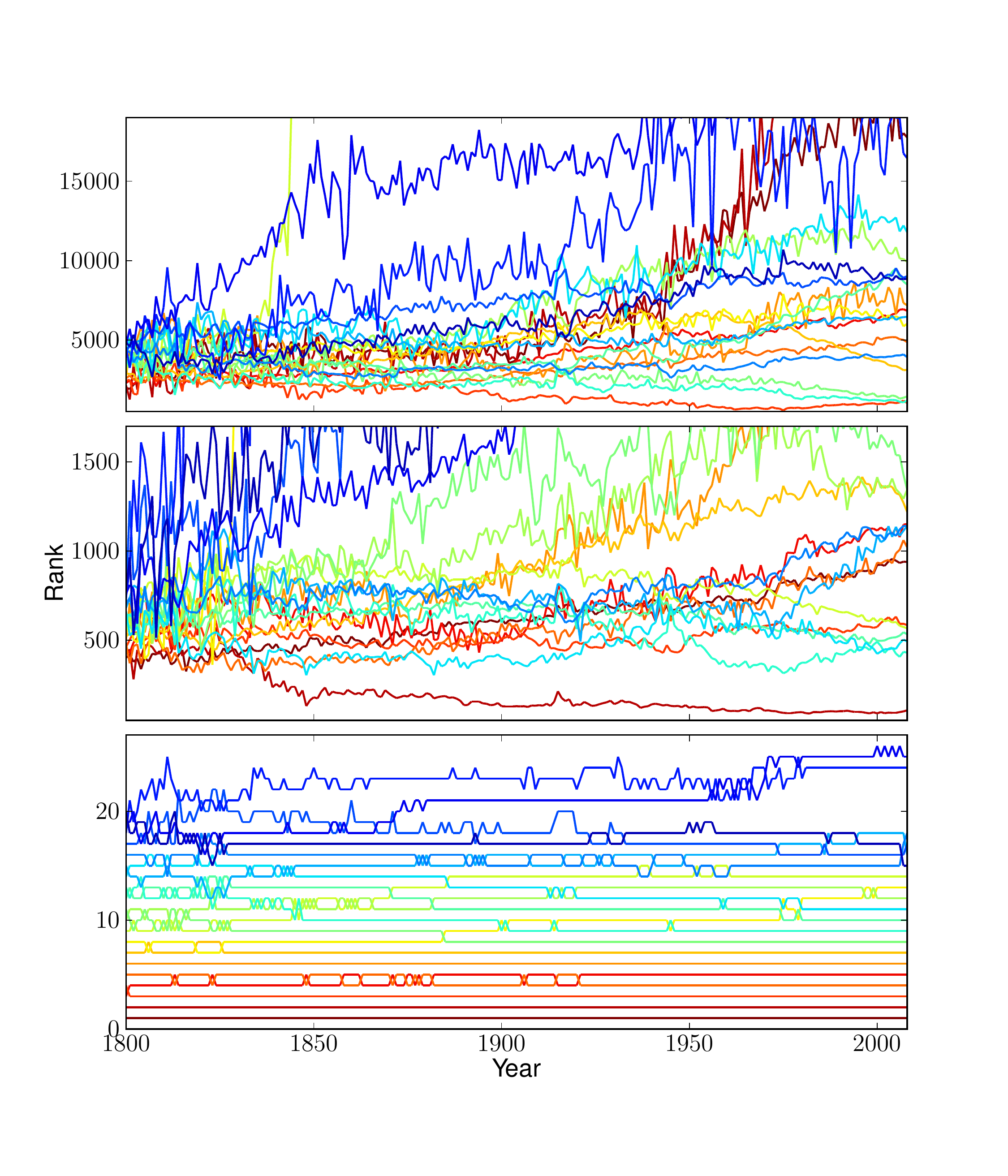}
     \caption{Rank variations in time of twenty words from three different scales for  French. }
  \label{fig:traject_fre}
\end{center}
\end{figure*} % }}}
\begin{figure*} % {{{ Espagueti
\begin{center}
     \includegraphics[width=.9\textwidth]{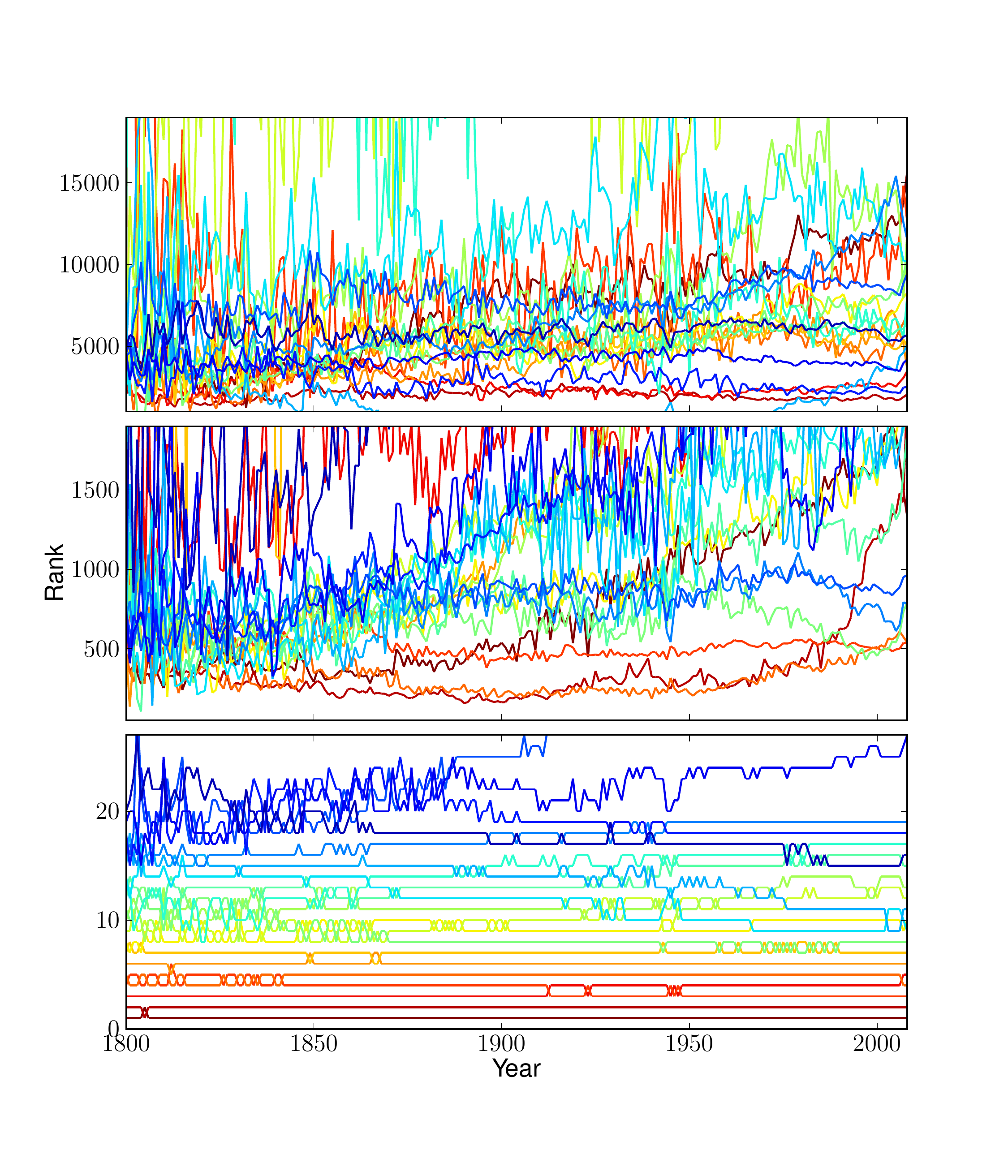}
     \caption{Rank variations in time of twenty words from three different scales for  Italian. }
  \label{fig:traject_ita}
\end{center}
\end{figure*} % }}}
\begin{figure*} % {{{ Espagueti
\begin{center}
     \includegraphics[width=.9\textwidth]{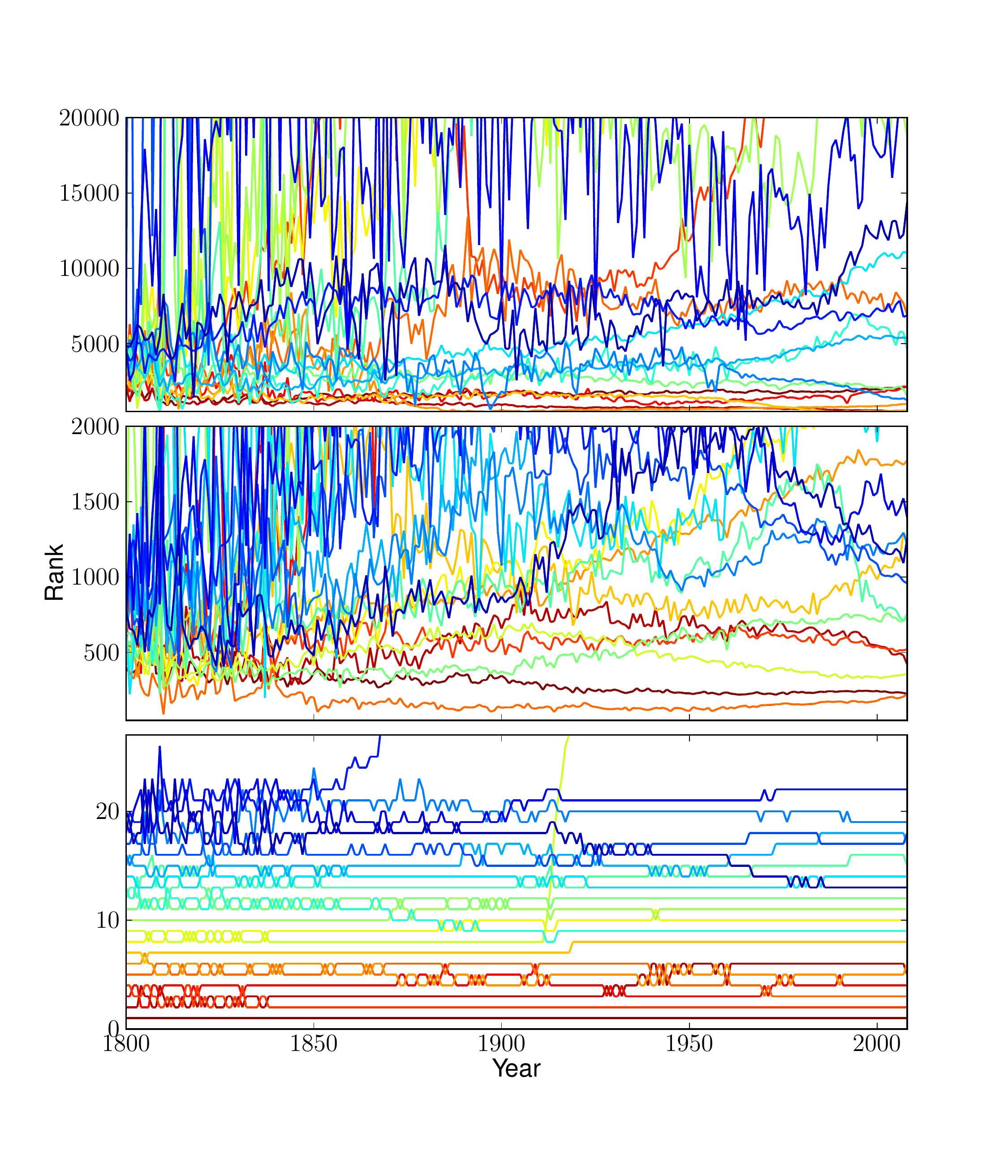}
     \caption{Rank variations in time of twenty words from three different scales for  Spanish. }
  \label{fig:traject_spa}
\end{center}
\end{figure*} % }}}
\begin{figure*} % {{{ Espagueti
\begin{center}
     \includegraphics[width=.9\textwidth]{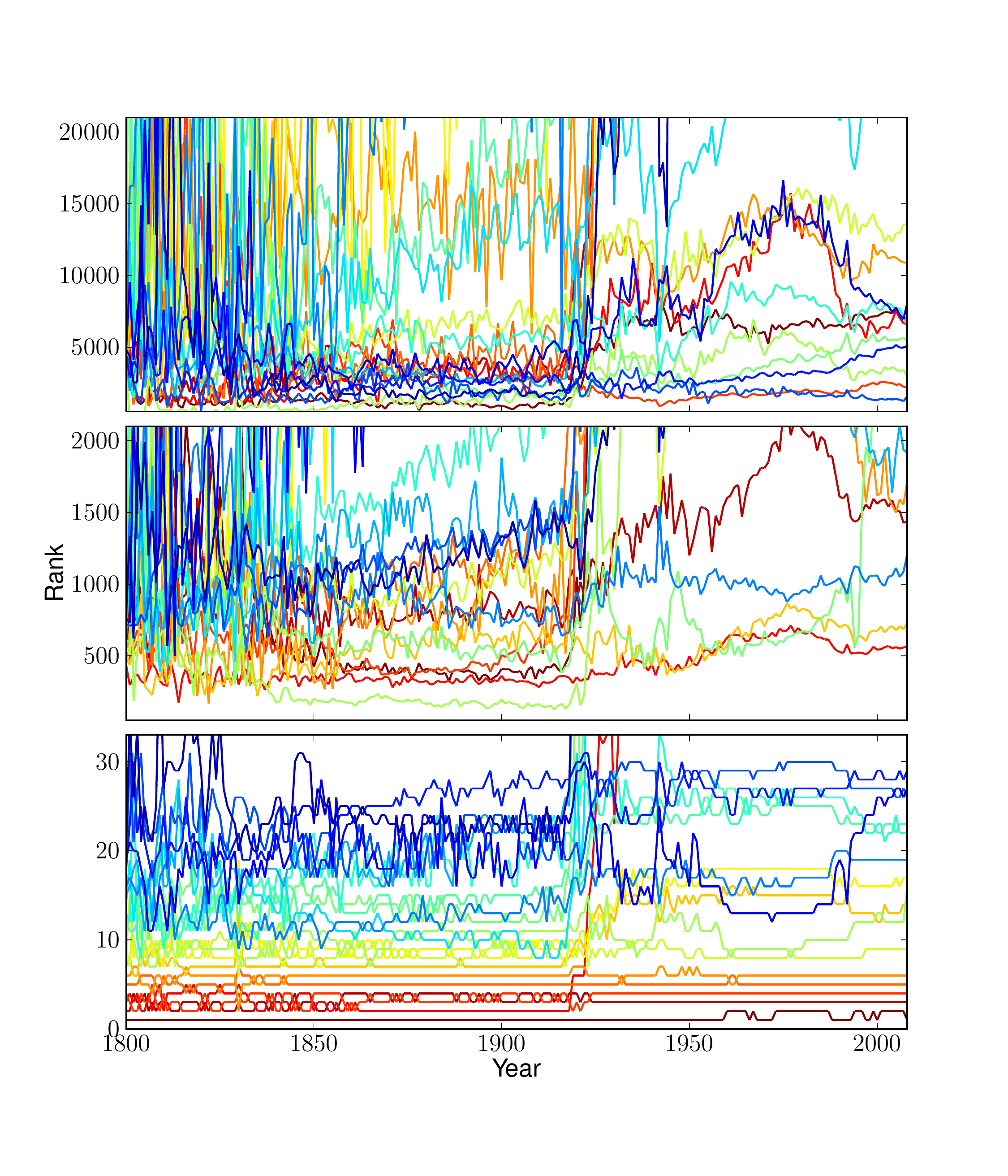}
     \caption{Rank variations in time of twenty words from three different scales for  Russian. }
  \label{fig:traject_rus}
\end{center}
\end{figure*} % }}}
\begin{figure*} % {{{ Espagueti
\begin{center}
     \includegraphics[width=.9\textwidth]{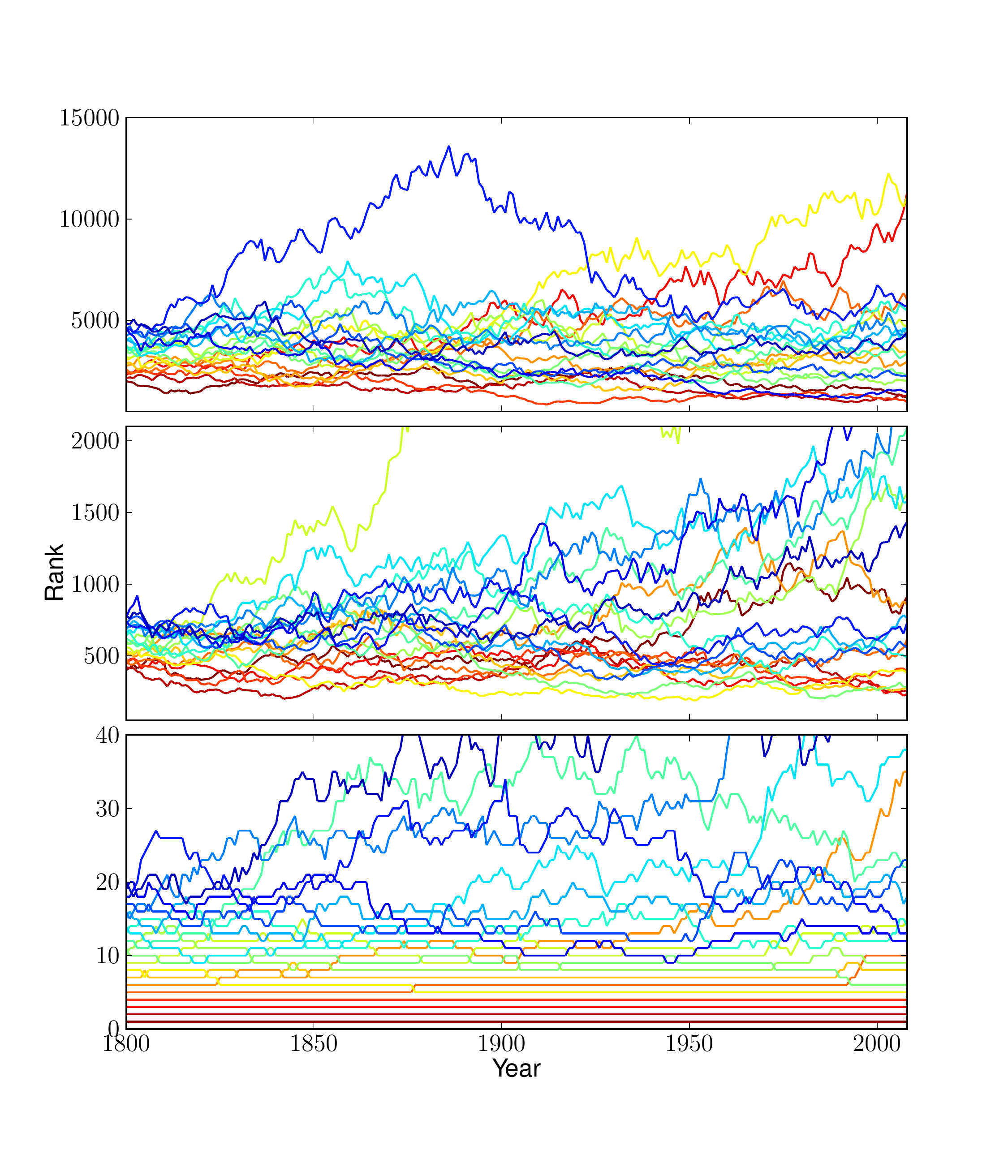}
     \caption{Rank variations in time of twenty words from three different scales for our simulated language. }
  \label{fig:traject_sim}
\end{center}
\end{figure*} % }}}

\textcolor{black}{Fig.~\ref{fig:allFlightDistributions} shows the distribution of relative flights for all languages. See main text for details.}

\begin{figure}[H] % {{{ All flight distributions
\includegraphics[width=\textwidth]{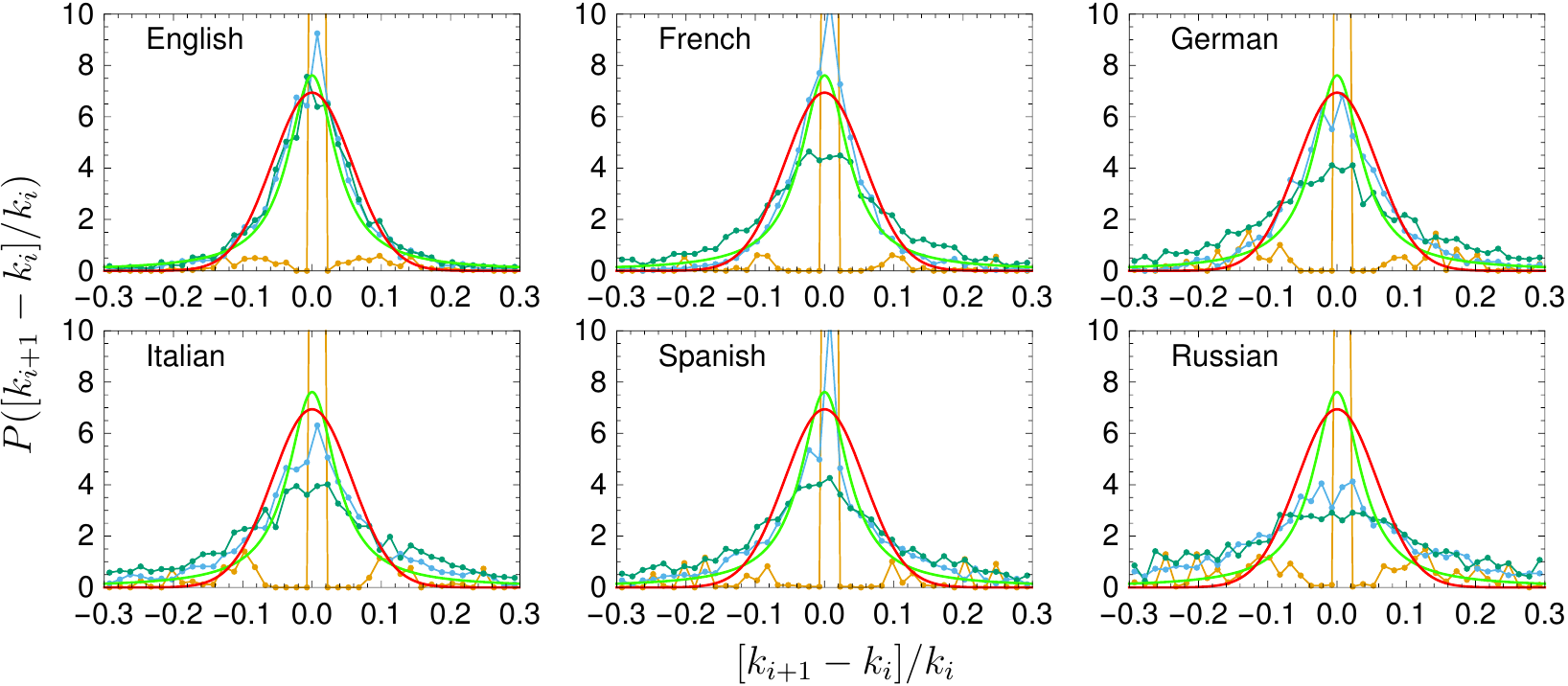}
\caption{\textcolor{black}{{\bf Distribution of relative flights for all languages studied.} 
A similar plot as the one presented in 
figure 6 is shown for other languages. The same color coding 
and details are used. 
}
}
\label{fig:allFlightDistributions}
\end{figure} % }}}

% }}}
\section{Correlation of relative frequency changes} % {{{
\textcolor{black}{We studied the correlations of the relative frequency changes (flights), defined in  the main text as
\begin{equation}
\Delta_t=\left(k_{t+1}-k_{t}\right)/k_{t}.
\end{equation}
We shall use a normalized version of it:
\begin{equation}
d_t = \frac{\Delta_t - \langle \Delta_t \rangle }{\sqrt{\langle (\Delta_t  -\langle \Delta_t \rangle )^2\rangle }},
\end{equation}
where $\langle \cdot \rangle$ denotes average over time. This normalization
ensures that both $\langle d_t \rangle = 0 $ and $\langle d_t ^2 \rangle = 1$.
% We now define the time correlation as 
The time correlation is given by
\begin{equation}
C_\tau = \langle d_t d_{t+\tau} \rangle.
\label{eq:correlation}
\end{equation}
In principle, this quantity also depends on $t$, but usually this dependence
is very weak, as in this case, and one can ignore it. }

\textcolor{black}{In Fig.~\ref{fig:word:walks:correlation} we show the average of $C_\tau$, of 50 different ranks chosen randomly, for
different
languages, as well as for the simulated language. 
We note that the correlation is very small, except for $\tau=0$, where it is
1, due to the normalization chosen, and for $\tau=1$ where a negative value, 
typical of bounded sequences, is observed for the six languages studied here. 
% Our model thus accounts for the correct behavior except at $\tau=1$, where
% some refinement could be introduced in
% This reinforces the random Gaussian model where the correlation coefficient
% is zero for all $\tau>0$. 
% 
% 
The random Gaussian model reproduces well these correlations except at $\tau=1$. 
}

% In the rthe correlation coefficient is zero for 
% all $\tau >0$, so 

\begin{figure}[H] % {{{ Word walks, correlations
\centering
         \includegraphics[width=.8\textwidth]{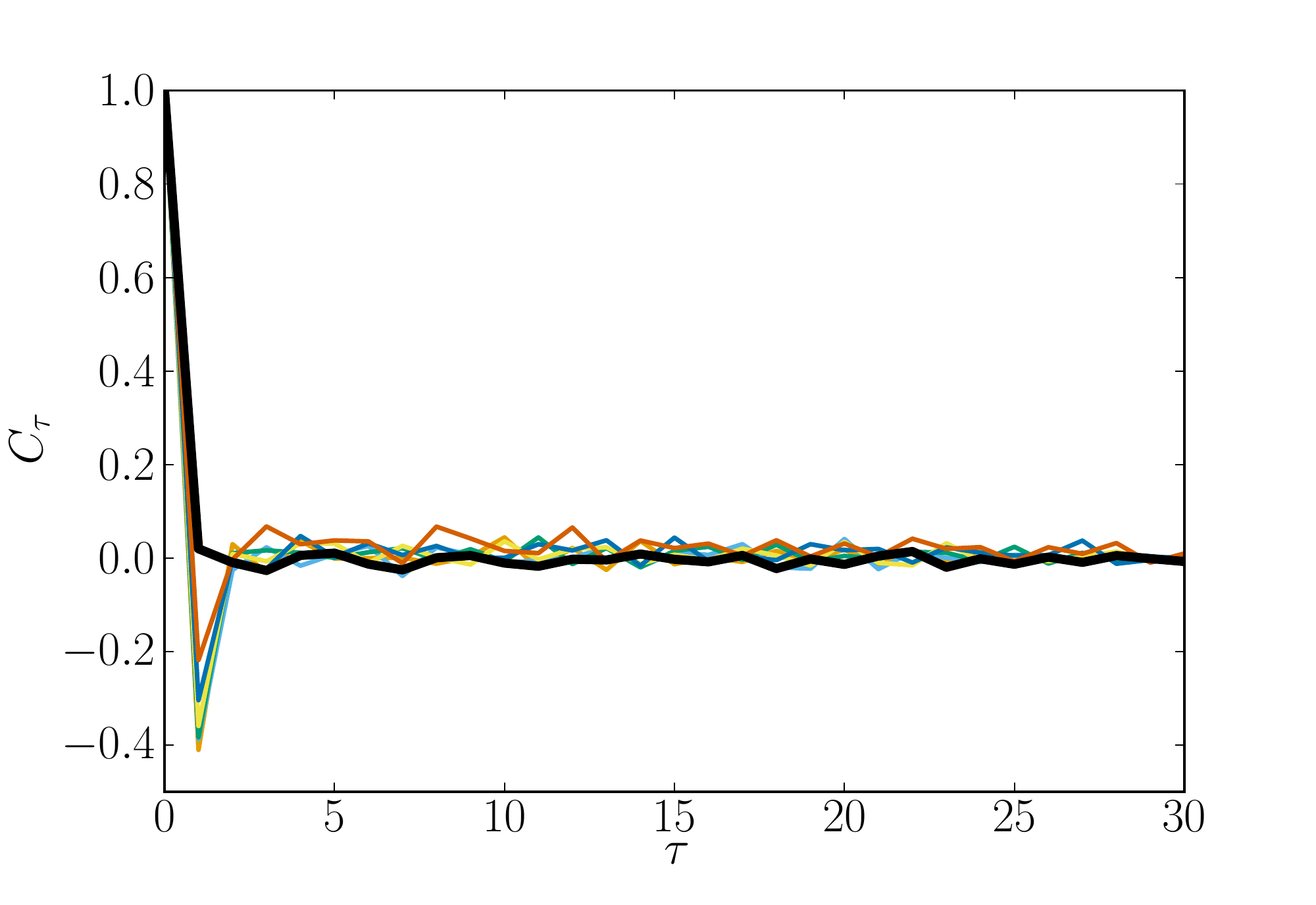}
     \caption{\textcolor{black}{{\bf Correlations for relative frequency changes} for different languages. Black line shows correlations for the simulated language.}}
  \label{fig:word:walks:correlation}
\end{figure} % }}}
% }}}
% }}}
\end{document}
\section*{Supporting Information Legends} % {{{
%
% Please enter your Supporting Information captions below in the following format:
%\item{\bf Figure SX. Enter mandatory title here.} Enter optional descriptive information here.
% 
%\begin{description}
%\item {\bf}
%\item {\bf}
%\end{description}
\setcounter{figure}{0}
\renewcommand{\thefigure}{S\arabic{figure}}
\renewcommand{\thesubfigure}{\thefigure\alph{subfigure}} 
\makeatletter 
	\renewcommand{\p@subfigure}{} 
	\renewcommand{\@thesubfigure}{\thesubfigure:\hskip\subfiglabelskip} 
\makeatother 
\begin{figure}[H] % {{{ 2000, several languages
     \subfigure{
          \label{fig:several:fixes:year}
          \includegraphics[width=.45\textwidth]{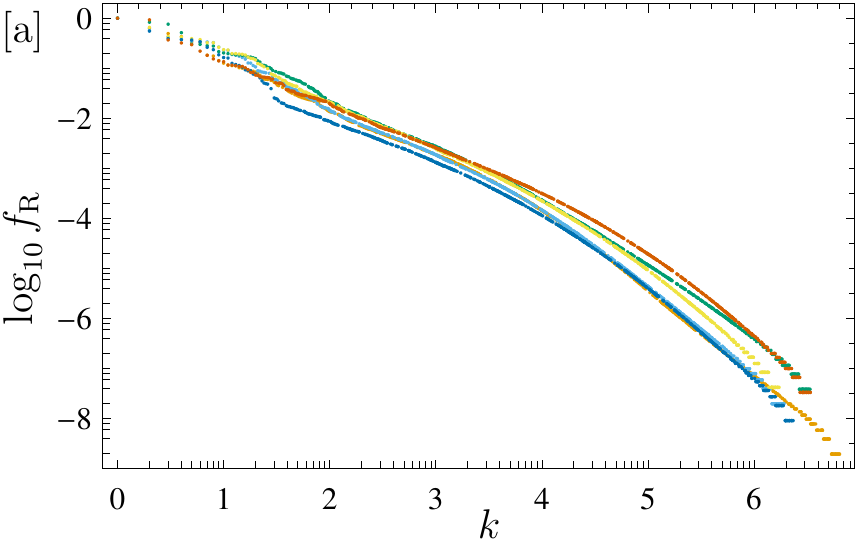}}
     \hspace{.3in}
      \subfigure{
           \label{fig:english:example}
           \includegraphics[width=.45\textwidth]{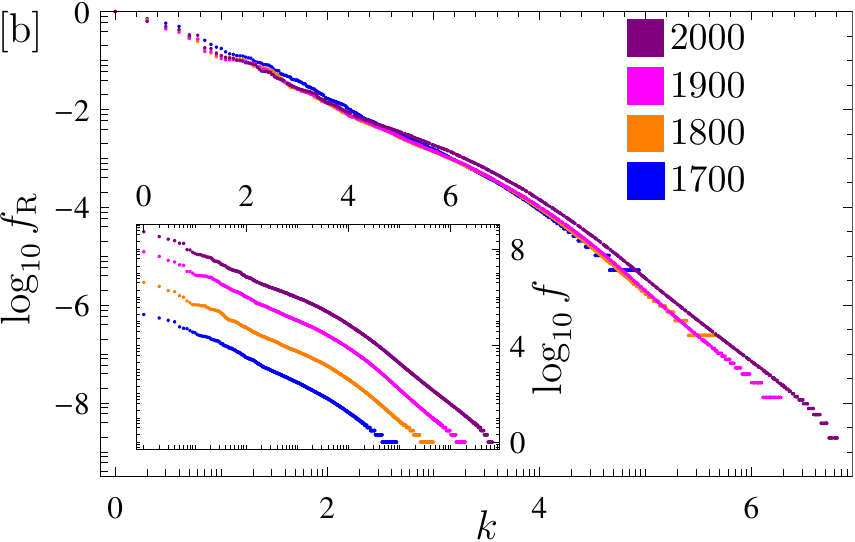}}
\caption{{\bf Rank distributions of words according to frequency.} [a]: Normalized
word frequency $f_\textup{R}$ as a function of the rank $k$ for several
languages for books published in the year 2000.
     The color code for languages is as follows: 
     {\color{colorb} $\blacksquare$} for French, 
     {\color{colorc} $\blacksquare$} for German, 
     {\color{colord} $\blacksquare$} for Italian, 
     {\color{colora} $\blacksquare$} for English,
     {\color{colore} $\blacksquare$} for Spanish, and
     {\color{colorf} $\blacksquare$} for Russian.
      [b]: Word frequency $f_\textup{R}$ as a function of
the rank $k$ for English and several years, normalized so that the most
frequent element has relative frequency one. In the inset, the unnormalized
frequency $f$ is shown.} \label{fig:Zipfians}
\end{figure} % }}}
\begin{figure}[H] % {{{ Fit
\includegraphics[width=\textwidth]{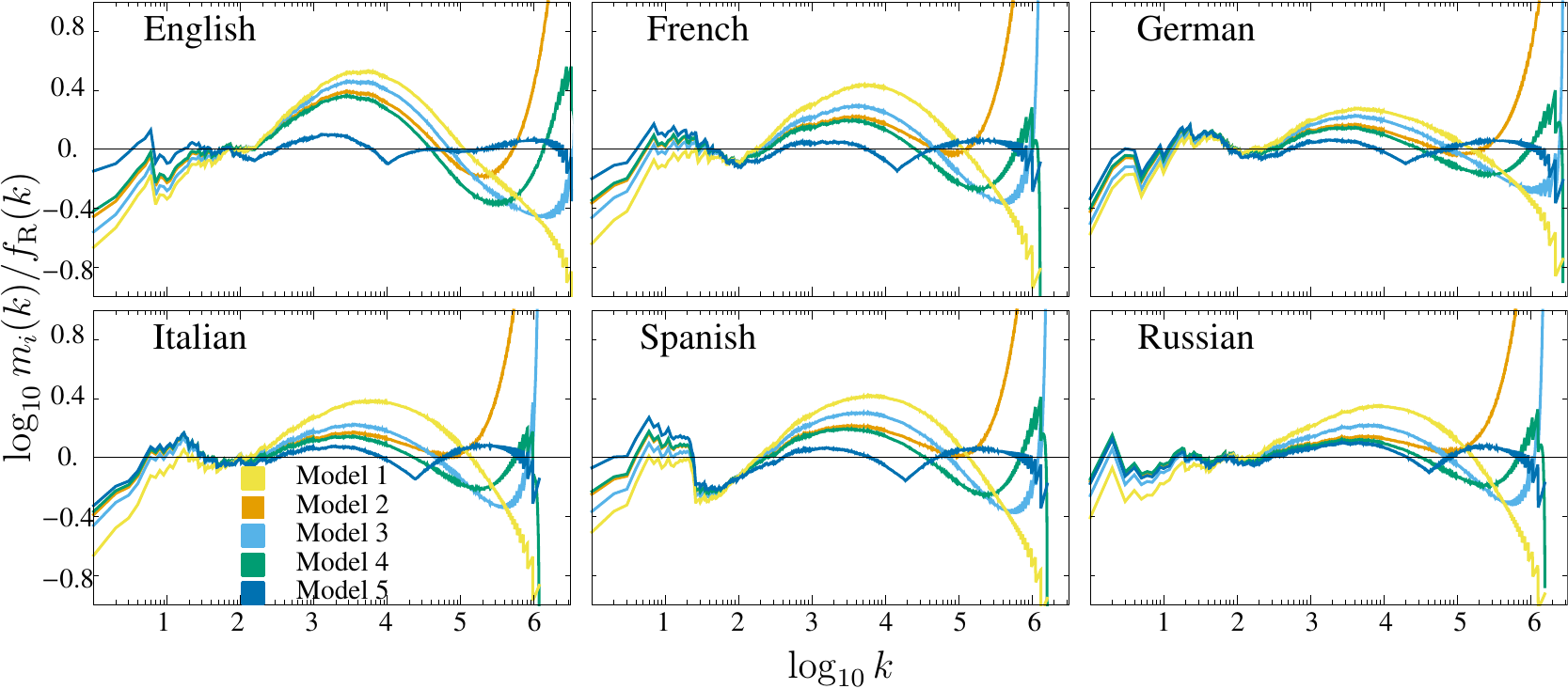}
\caption{{\bf Comparison between the different models,
equations S1--S5, and the frequency of rank distribution.} 
We use the data for the year 2000 and all languages under
consideration.  The logarithm base 10 of the ratio of the observed
values and the model is plotted.  It can be appreciated that different models fit better in
different regions. However there is no model that fits all languages and all
regions much better than the others. }
\label{fig:fits}
\end{figure} % }}}
\begin{figure}[H] % {{{ Espagueti
\begin{center}
     \includegraphics[width=.9\textwidth]{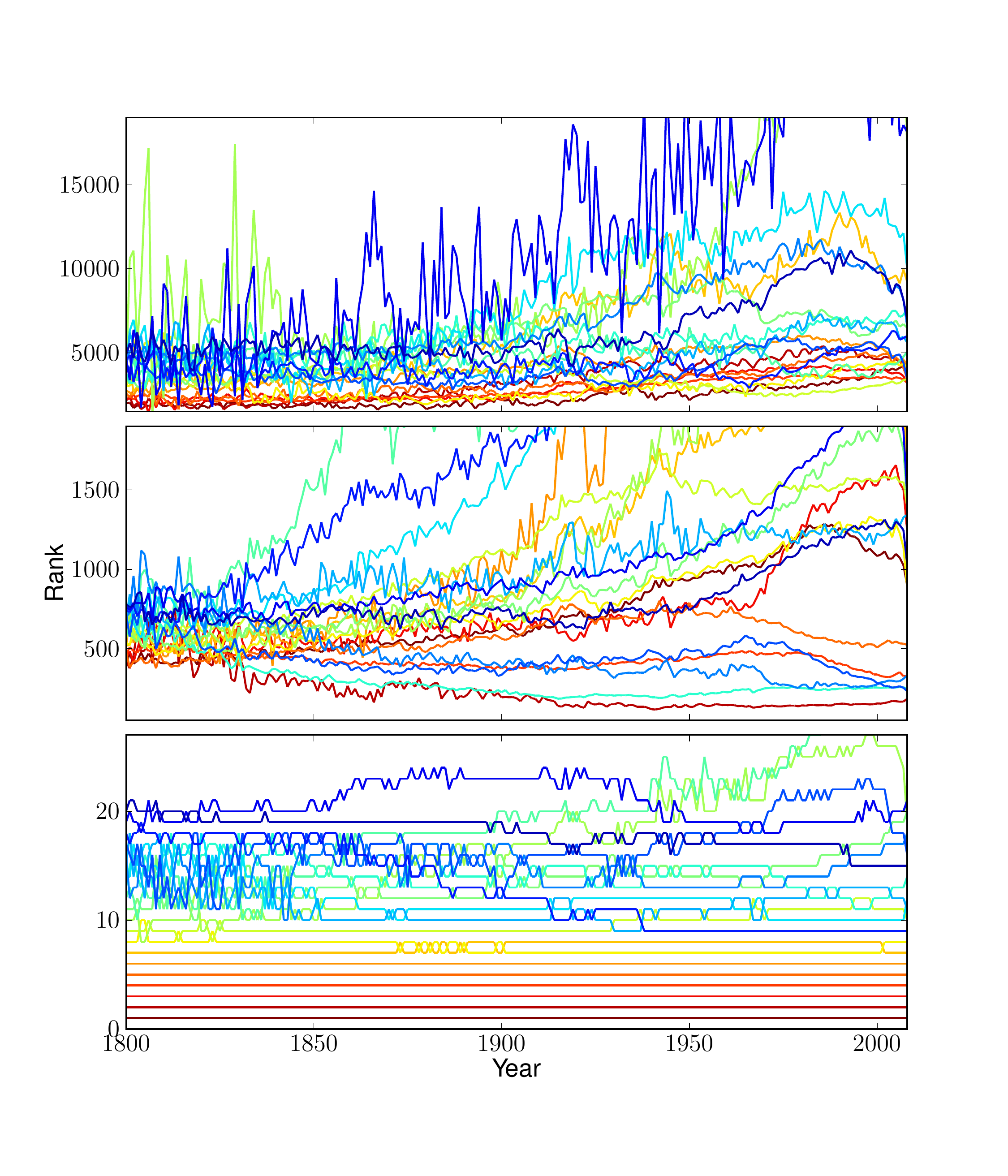}
     \caption{{\bf Rank variations in time of twenty words from three different scales for  English. }}
  \label{fig:traject_eng}
\end{center}
\end{figure} % }}}
\begin{figure}[H] % {{{ Espagueti
\begin{center}
     \includegraphics[width=.9\textwidth]{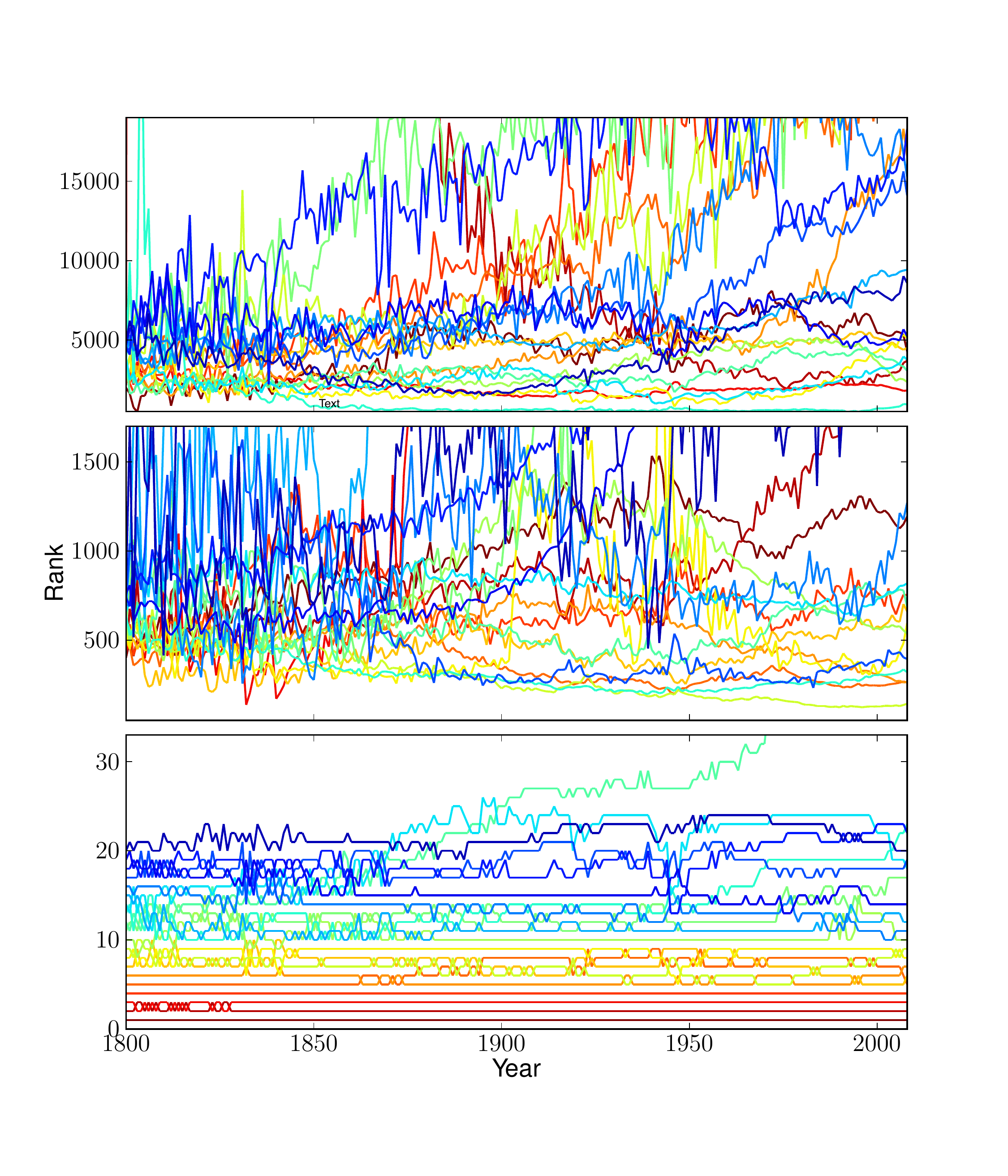}
     \caption{{\bf Rank variations in time of twenty words from three different scales for  German. }}
  \label{fig:traject_ger}
\end{center}
\end{figure} % }}}
\begin{figure}[H] % {{{ Espagueti
\begin{center}
     \includegraphics[width=.9\textwidth]{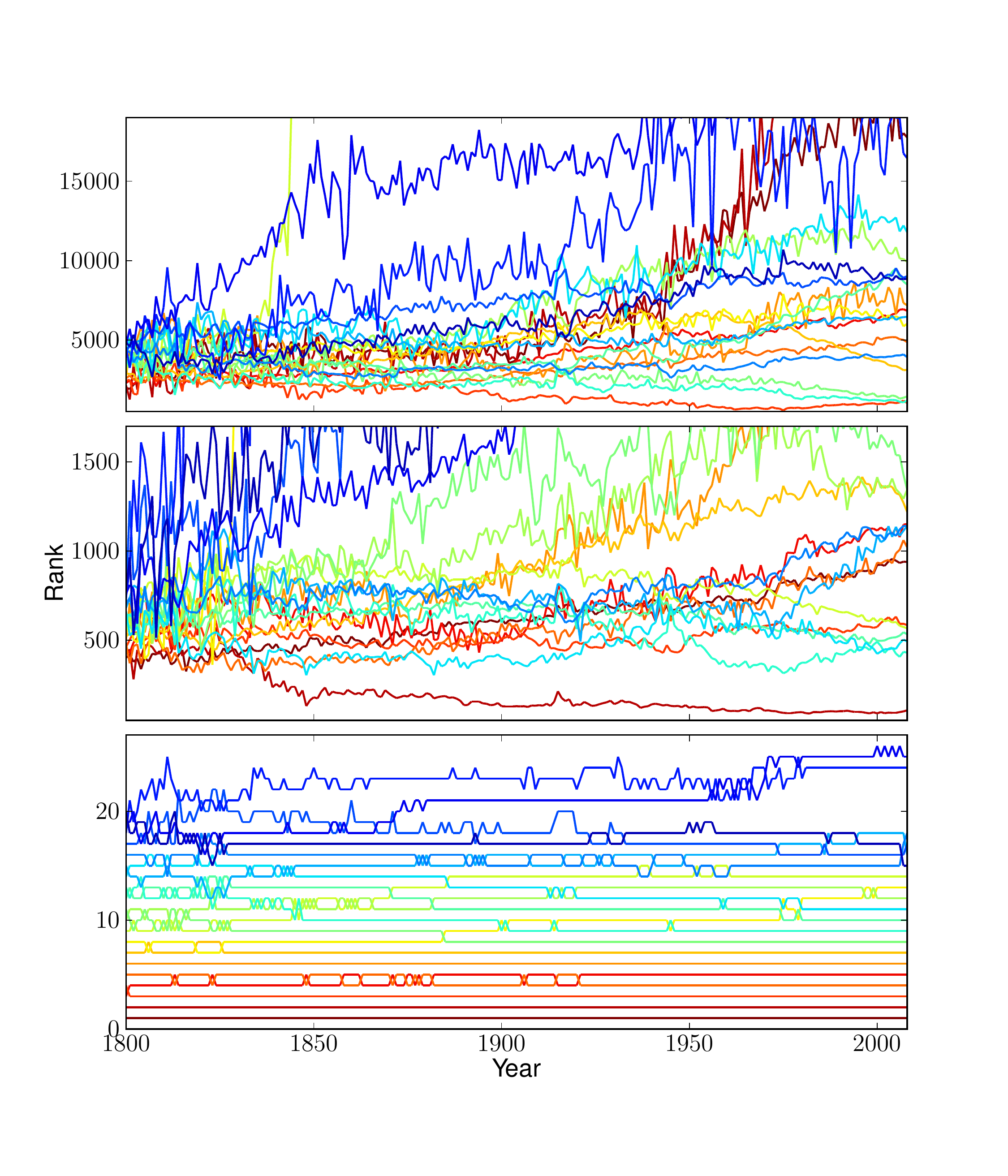}
     \caption{{\bf Rank variations in time of twenty words from three different scales for  French. }}
  \label{fig:traject_fre}
\end{center}
\end{figure} % }}}
\begin{figure}[H] % {{{ Espagueti
\begin{center}
     \includegraphics[width=.9\textwidth]{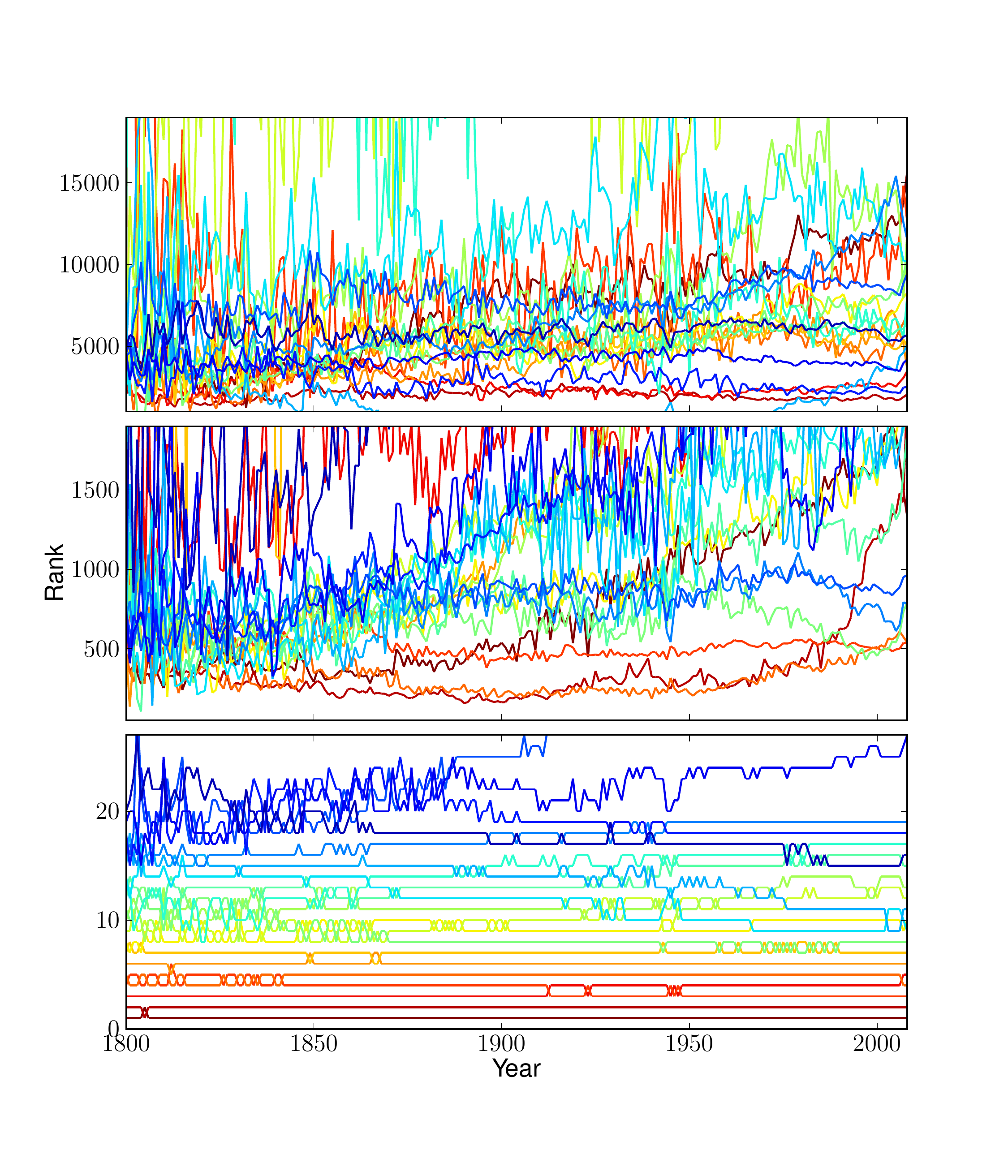}
     \caption{{\bf Rank variations in time of twenty words from three different scales for  Italian. }}
  \label{fig:traject_ita}
\end{center}
\end{figure} % }}}
\begin{figure}[H] % {{{ Espagueti
\begin{center}
     \includegraphics[width=.9\textwidth]{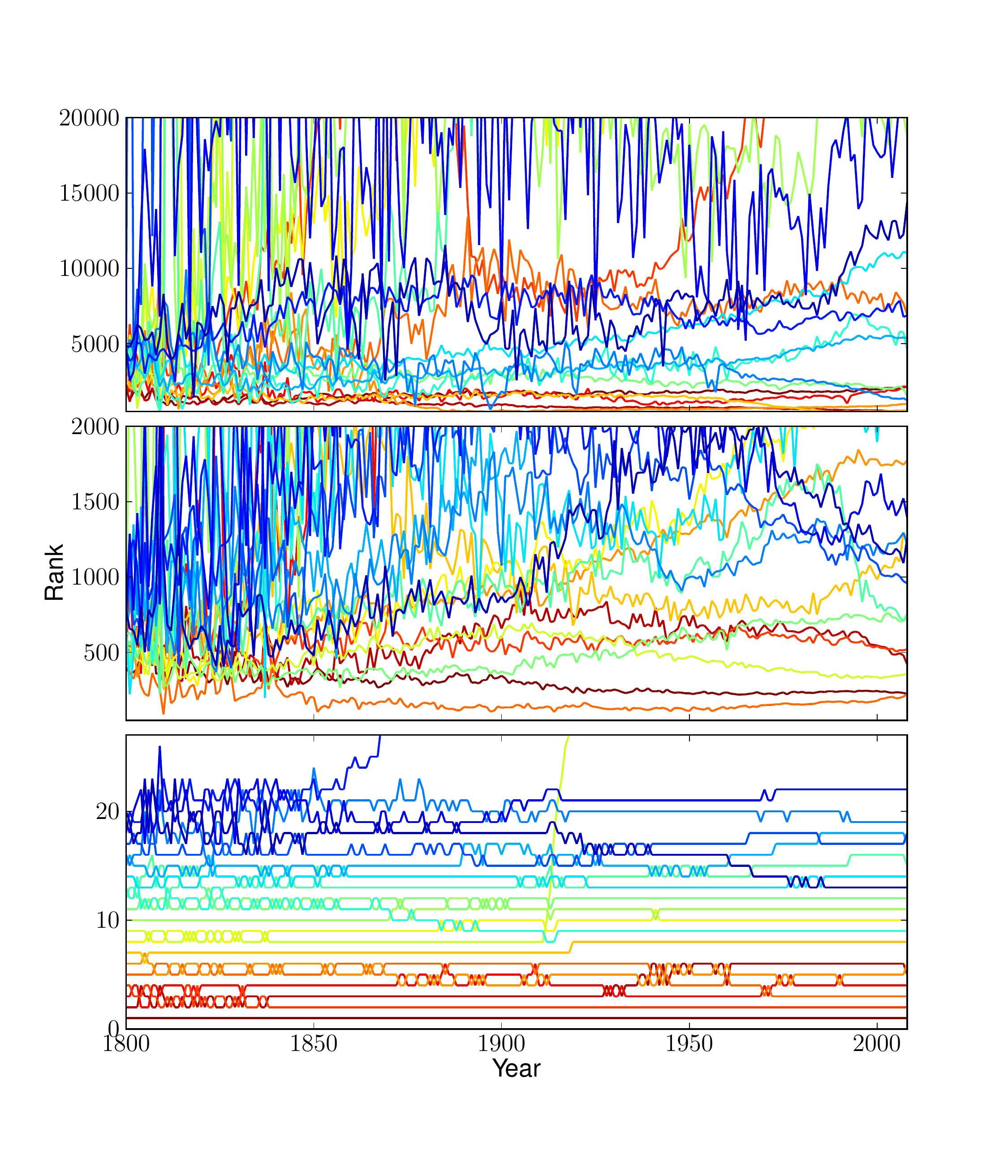}
     \caption{{\bf Rank variations in time of twenty words from three different scales for  Spanish. }}
  \label{fig:traject_spa}
\end{center}
\end{figure} % }}}
\begin{figure}[H] % {{{ Espagueti
\begin{center}
     \includegraphics[width=.9\textwidth]{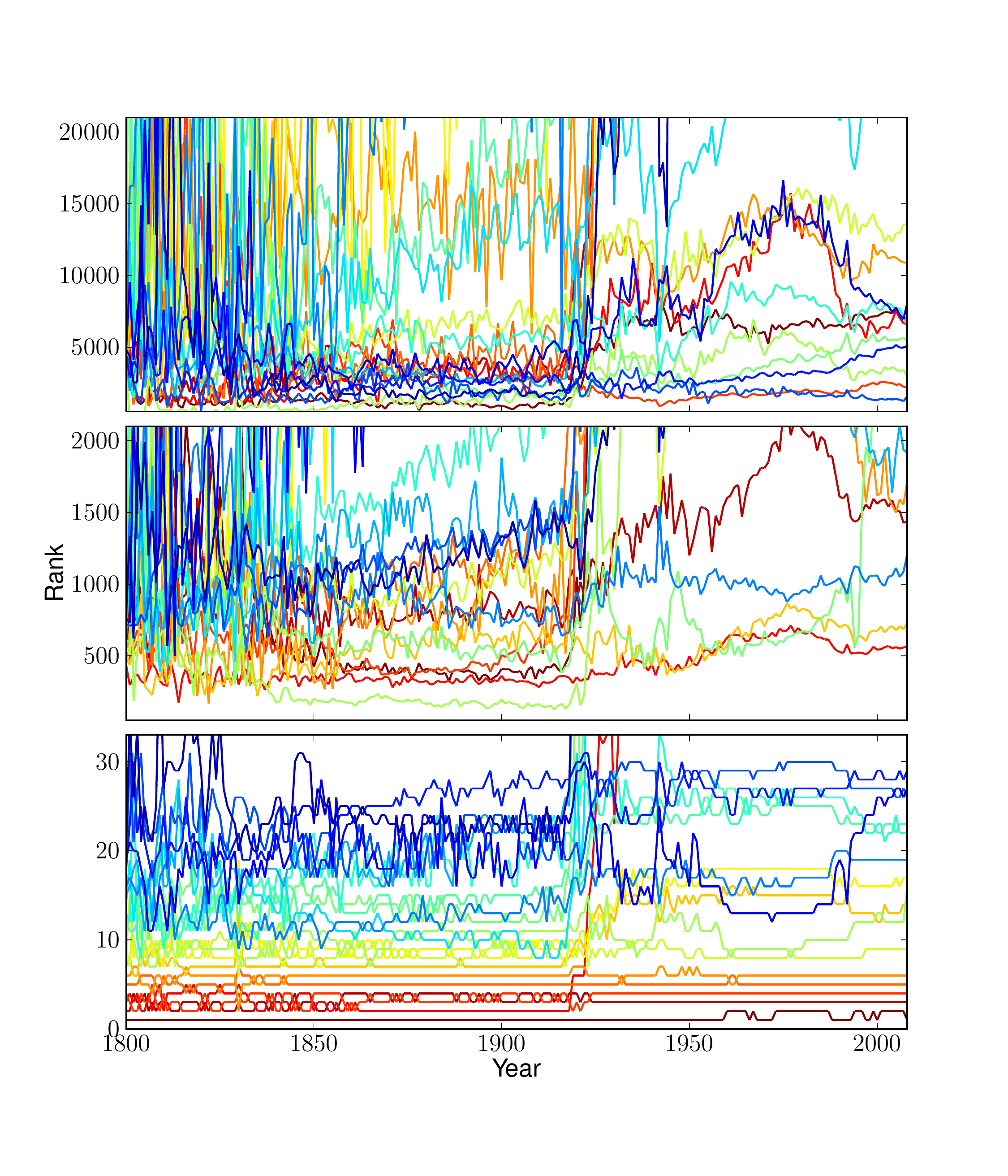}
     \caption{{\bf Rank variations in time of twenty words from three different scales for  Russian. }}
  \label{fig:traject_rus}
\end{center}
\end{figure} % }}}
\begin{figure}[H] % {{{ Espagueti
\begin{center}
     \includegraphics[width=.9\textwidth]{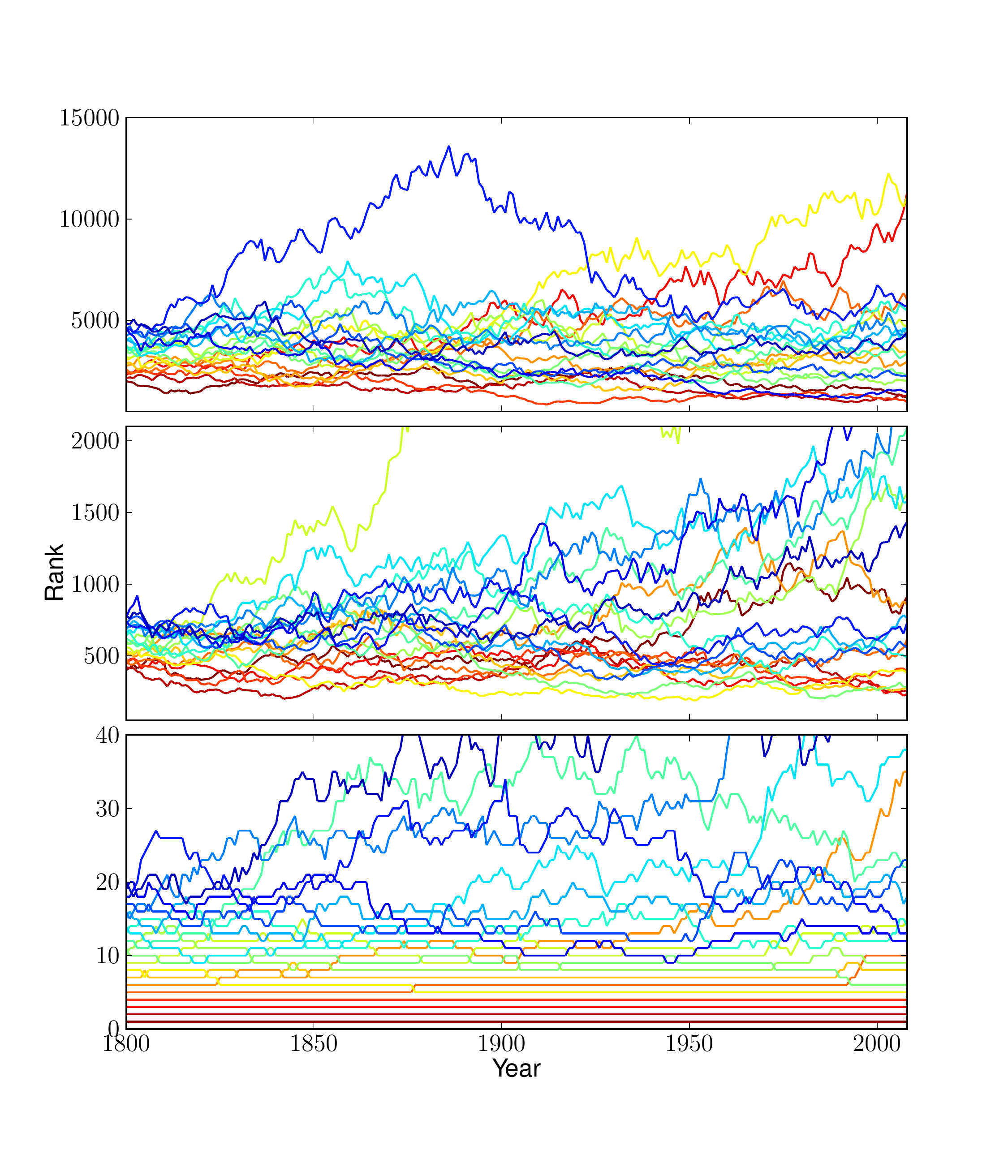}
     \caption{{\bf Rank variations in time of twenty words from three different scales for our \textcolor{black}{simulated} language. }}
  \label{fig:traject_sim}
\end{center}
\end{figure} % }}}
\begin{figure}[H] % {{{ All flight distributions
\includegraphics[width=\textwidth]{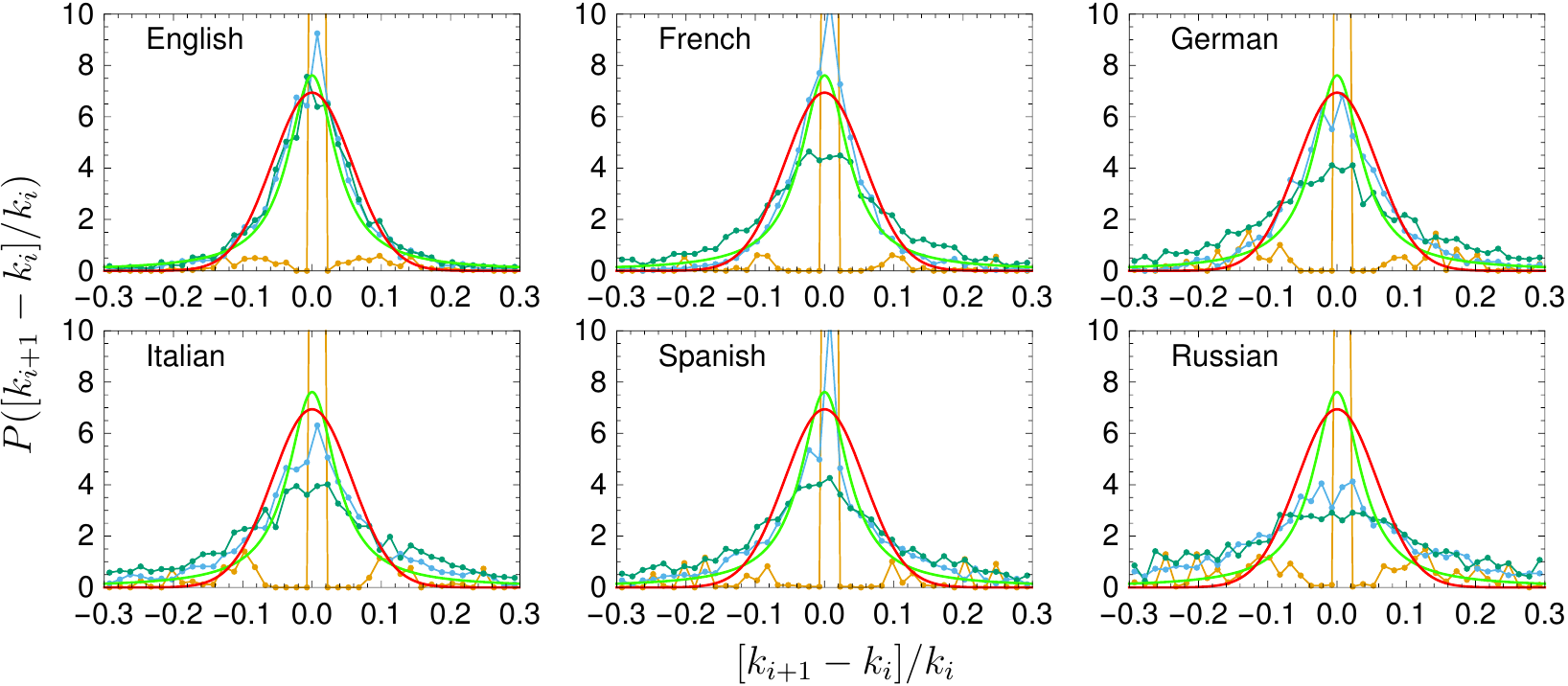}
\caption{\textcolor{black}{{\bf Distribution of relative flights for all languages studied.} 
A similar plot as the one presented in 
\fref{fig:flight:distribution} is shown for other languages. The same color coding 
and details are used. 
}
}
\label{fig:allFlightDistributions}
\end{figure} % }}}
\begin{figure}[H] % {{{ Word walks, correlations
\centering
         \includegraphics[width=.8\textwidth]{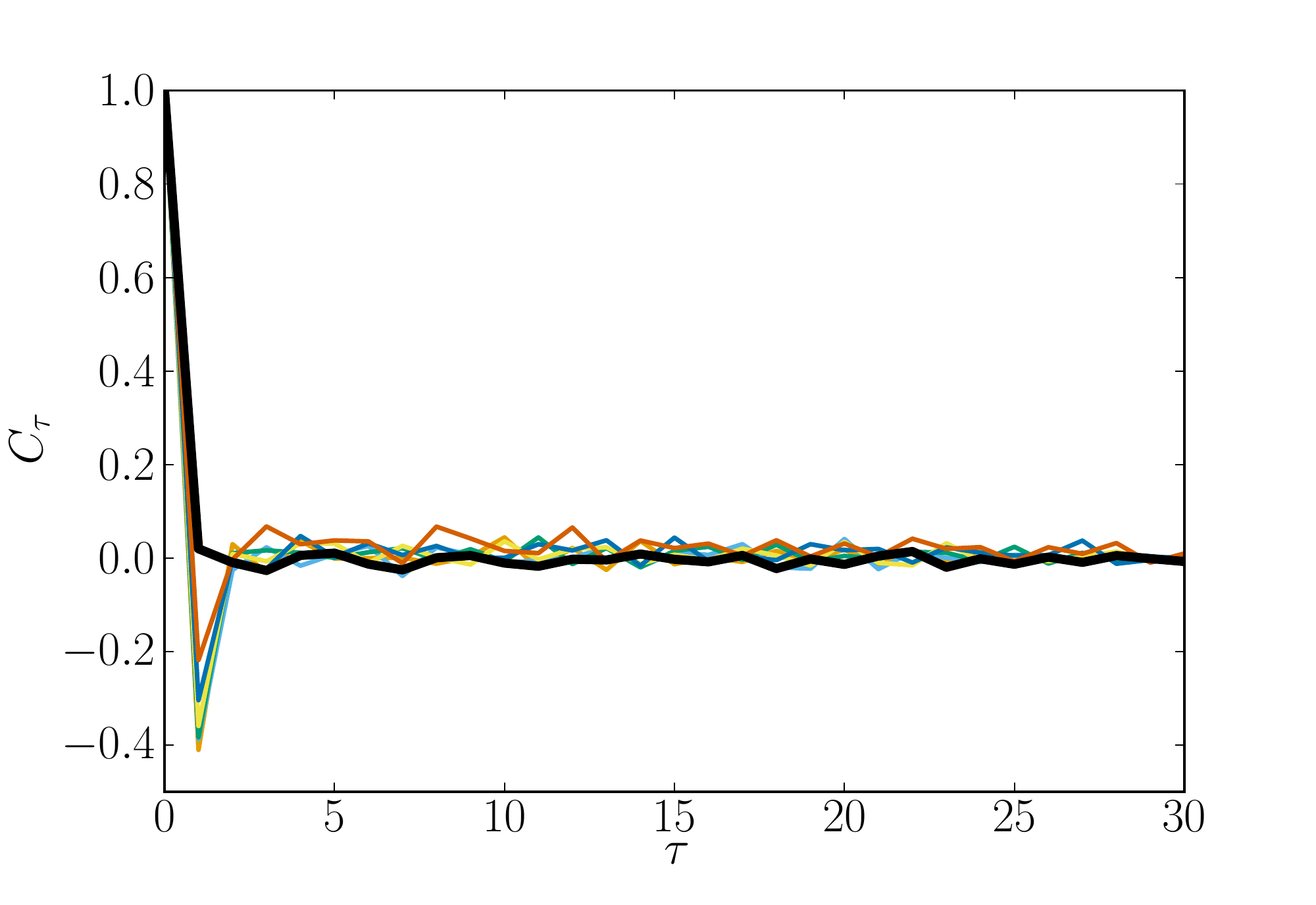}
     \caption{\textcolor{black}{{\bf Correlations for relative frequency changes} for different languages. Black line shows correlations for simulated language.}}
  \label{fig:word:walks:correlation}
\end{figure} % }}}
% }}}